\documentclass[10pt,twocolumn,letterpaper]{article}

\usepackage{iccv}
\usepackage{times}
\usepackage{epsfig}
\usepackage{graphicx}
\usepackage{amsmath}
\usepackage{amssymb}
\usepackage{multirow}
\usepackage{booktabs}
\usepackage{lettrine}
\usepackage{color}
\usepackage{subcaption}
\usepackage{url}
\usepackage{verbatim}
\usepackage{array}
\usepackage{stix}
\usepackage{microtype}
\usepackage[english]{babel}
\usepackage[utf8]{inputenc}
\usepackage{amsfonts}
\usepackage[ruled,linesnumbered]{algorithm2e}
\usepackage[noend]{algpseudocode}
\usepackage{enumerate}
\usepackage{xcolor}
\usepackage{xspace}
\usepackage[subtle,mathspacing=normal,mathdisplays=tight,tracking=normal]{savetrees}

\SetCommentSty{mycommfont}

\newcommand{\todo}[1]{\textcolor[rgb]{1,0,0}{(#1)}}
\newcommand{\cut}[1]{}
\newcommand{\keypoint}[1]{\vspace{0.1cm}\noindent\textbf{#1}\quad}

\newcommand{\NIW}{\mathcal{NIW}}
\newcommand{\Gauss}{\mathcal{N}}
\newcommand{\T}{\mathcal{T}}
\DeclareMathOperator*{\argmin}{arg\,min}
\DeclareMathOperator*{\argmax}{arg\,max}

\def\tierIN{\textit{tiered}ImageNet}
\def\miniIN{\textit{mini}ImageNet}

\usepackage{color, colortbl}
\definecolor{Gray}{gray}{0.9}

\usepackage[pagebackref=true,breaklinks=true,letterpaper=true,colorlinks,bookmarks=false]{hyperref}

\iccvfinalcopy 

\ificcvfinal\fi

\begin{document}

\title{Shallow Bayesian Meta Learning for Real-World Few-Shot Recognition}

\author{Xueting Zhang$^{1}$\thanks{Xueting and Debin contributed equally to this research, code is available \url{https://github.com/Open-Debin/Bayesian_MQDA} }, 
Debin Meng$^{3*}$, Henry Gouk$^{1}$, Timothy Hospedales$^{1,2}$
\\
$^{1}$University of Edinburgh, Edinburgh, United Kingdom\\
$^{2}$Samsung AI Centre, Cambridge, United Kingdom\\
$^{3}$University of Chinese Academy of Sciences, Beijing, China \\
{\tt\small \{xueting.zhang,henry.gouk,t.hospedales\}@ed.ac.uk, mengdebin16@mails.ucas.ac.cn}
}

\maketitle
\ificcvfinal\thispagestyle{empty}\fi

\begin{abstract}
Many state-of-the-art few-shot learners focus on developing effective training procedures for feature representations, before using simple (e.g., nearest centroid) classifiers. We take an approach that is agnostic to the features used, and focus exclusively on meta-learning the final classifier layer. Specifically, we introduce MetaQDA, a Bayesian meta-learning generalisation of the classic quadratic discriminant analysis. This approach has several benefits of interest to practitioners: meta-learning is fast and memory efficient, without the need to fine-tune features. It is agnostic to the off-the-shelf features chosen, and thus will continue to benefit from future advances in feature representations. Empirically, it leads to excellent performance in cross-domain few-shot learning, class-incremental few-shot learning, and crucially for real-world applications, the Bayesian formulation leads to state-of-the-art uncertainty calibration in predictions.
\end{abstract}

\section{Introduction}

Few-shot recognition methods aim to solve classification problems with limited labelled training data, motivating a large body of work \cite{wang2019fewShotSurvey}. Contemporary approaches to few-shot recognition are characterized by a focus on deep meta-learning \cite{hospedales2020metaSurvey} methods that provide data efficient learning of new categories by using auxiliary data to train a model designed for rapid adaptation to new categories \cite{finn2017model,zintgraf2018cavia}, or for synthesizing a classifier for new categories in a feed-forward manner \cite{mishra2018simple, qiao2017few}. Most of these meta-learning methods have been intimately interwoven with the training algorithm and/or architecture of the deep network that they build upon. For example, many have relied on episodic training schemes \cite{snell2017prototypical,vinyals2016matching}, where few-shot learning problems are simulated at each iteration of training; differentiable optimisers \cite{bertinetto2019R2D2,lee2019meta}, or new neural network modules \cite{sung2018learning,gordon2019metaPred} to facilitate data efficient learning and recognition.  

Against this backdrop, a handful of recent studies \cite{wang2019simpleshot,goldblum2020unraveling,chen2019closerfewshot,mangla2020charting,yin2020metaMemorisation,wang2020comparison} have pushed back against deep meta-learning. They have observed, for example, that a well tuned convolutional network pre-trained for multi-class recognition and combined with a simple linear or nearest centroid classifier can match or outperform state-of-the-art meta-learners. Even self-supervised pre-training \cite{mangla2020charting} has led to feature extractors that outperform many meta-learners. These analyses raise the question: \emph{is meta-learning indeed beneficial, or is focusing on improving conventional pre-training sufficient?}

We take a position in defense of meta-learning for few-shot recognition. To disentangle the influences of meta-learning per-se and feature learning discussed above, we restrict ourselves to \emph{fixed pre-trained features} and conduct no feature learning in this study. It shows that meta-learning, even in its shallowest form, can boost few-shot learning above and beyond whatever is provided by the pre-trained features alone. 

We take an amortized Bayesian inference approach \cite{gordon2019metaPred,heskes2000empirical} to shallow meta-learning. During meta-testing, we infer a distribution over classifier parameters given the support set; and during meta-training we learn a feed-forward inference procedure for these parameters. While the limited recent work in Bayesian meta-learning is underpinned by amortized Variational Inference \cite{gordon2019metaPred}, our approach relies instead on conjugacy \cite{gelman2003bda}. Specifically, we build upon the classic Quadratic Discriminant Analysis (QDA) \cite{friedman2001elements} classifier and extended it with a Bayesian prior, an inference pipeline for the QDA parameter posterior given the support set, and  gradient-based meta-training. We term the overall framework \textit{MetaQDA}. 

MetaQDA has several practical benefits for real-world deployments. Firstly, MetaQDA allows meta-learning to be conducted in the resource constrained scenario without end-to-end training \cite{ignatov2019aiSmartphone}, while providing superior performance to fixed-feature approaches \cite{wang2019simpleshot,chen2019closerfewshot,mangla2020charting}. Furthermore by decomposing representation learning from classifier meta-learning, MetaQDA is expected to benefit from continued progress in CNN architectures and training strategies. Indeed our empirical results show our feature-agnostic strategy benefits a diverse range of classic and recent feature representations.

Secondly, as computer vision systems begin to be deployed in high-consequence applications where safety  \cite{Kuper2018TowardNetworks}
or fair societal outcomes \cite{Du2019FairnessPerspective} are at stake, their \emph{calibration} becomes as equally, or more, important as their actual accuracy. E.g., Models must reliably report low-certainty in those cases where they do make mistakes, thus allowing their decisions in those cases to be reviewed. Indeed, proper calibration is a hard requirement for deployment in many high importance applications \cite{guo2017calibration,nixon2019measuring}. Crucially, we show that our Bayesian MetaQDA leads to significantly better calibrated models than the standard classifiers in the literature. 

Finally, we show that MetaQDA has particularly good performance in {cross-domain} scenarios where existing methods are weak \cite{chen2019closerfewshot}, but which are ubiquitious in practical applications, where there is invariably insufficient domain-specific data to conduct in-domain meta-learning \cite{guo2020boarder}. Furthermore, as a Bayesian formulation, MetaQDA is inherently suited to the highly practical, but otherwise hard to achieve setting of \emph{incremental} \cite{tao2020few,rebuffi2017icarl} few-shot learning, where it achieves state of the art performance `out of the box'.

To summarize our contributions: (i) We present MetaQDA, a novel and efficient Bayesian approach to classifier meta-learning based on conjugacy. (ii) We empirically demonstrate that MetaQDA's efficient fixed feature learning provides excellent performance across a variety of settings and metrics including conventional, cross-domain, class-incremental, and probability calibrated few-shot learning. (iii) We shed light on the meta-learning vs vanilla pre-training debate by disentangling the two and showing a clear benefit from meta-learning, across a variety of fixed feature representations.

\section{Related Work}
\keypoint{Few-Shot and Meta-Learning Overview}
Few-shot and meta-learning are now a widely studied area that is too broad to review here. We refer the reader to comprehensive recent surveys for an introduction and review \cite{wang2019fewShotSurvey,hospedales2020metaSurvey}. In general they proceed in two stages: meta-training the strategy for few-shot learning based on one or more auxiliary datasets; and meta-testing (learning new categories) on a target dataset, which should be done data-efficiently given the knowledge from meta-training.
A high level categorization of common approaches groups them into methods that (1) meta-learn how to perform rapid gradient-based adaptation during meta-test \cite{finn2017model,zintgraf2018cavia}; and (2) meta-learn a feed-forward procedure to synthesize a classifier for novel categories given an embedding of the support set \cite{gordon2019metaPred,qiao2017few}, where metric-based learners are included in the latter category \cite{hospedales2020metaSurvey}. 

\keypoint{Is Meta-Learning Necessary?} Many recent papers have questioned whether elaborate meta-learning procedures are necessary. 
SimpleShot \cite{wang2019simpleshot} observes vanilla CNN features pre-trained for recognition achieve near SotA performance when appropriately normalized and used in a trivial nearest centroid classifier (NCC). Chen et al.~\cite{chen2019closerfewshot} present the simple but high-performance Baseline++, based on fixing a pre-trained feature extractor and then building a linear classifier during meta-test.
\cite{goldblum2020unraveling} observe that although SotA meta-learned deep features do exhibit strong performance in few-shot learning, this feature quality can be replicated by adding simple compactness regularisers to vanilla classifier pre-training. S2M2 \cite{mangla2020charting} demonstrates that after pre-training a network with self-supervised learning and/or manifold-regularised vanilla classification, excellent few-shot recognition is achieved by simply training a linear classifier on the resulting representation. \cite{yin2020metaMemorisation} analyzes whether the famous MAML algorithm is truly meta-learning, or simply pre-training a strong feature.

We show that for fixed features pre-trained by several of the aforementioned ``off-the-shelf'' non-meta techniques \cite{wang2019simpleshot,mangla2020charting}, meta-learning \emph{solely} in classifier-space further improves performance. This allows us to  conclude that meta-learning \emph{does} add value, since alternative vanilla (i.e., non-meta) pre-training approaches do not influence the final classifier. We leave conclusive analysis of the relative merits of meta-learning vs vanilla pre-training of feature representation space to future work. In terms of empirical performance, we surpass all existing strategies based on fixed pre-trained features, and most alternatives based on deep feature meta-learning.

\keypoint{Fixed Feature Meta-Learning} A minority of meta-learning studies such as \cite{rusu2019leo,liu2020universal} have also built on fixed features. LEO \cite{rusu2019leo} synthesizes a classifier layer for a fixed feature extractor using a hybrid gradient- and feedforward-strategy. The concurrent URT \cite{liu2020universal} addresses multi-domain few-shot learning by meta-training a module that fuses an array of fixed features and dynamically produces a new feature encoding for a new domain. Ultimately, URT uses a ProtoNet  \cite{snell2017prototypical} classifier, and thus our contribution is orthogonal to URT's, as MetaQDA aims to replace the classifier (ie, ProtoNet), not produce a new feature. Indeed we show empirically that MetaQDA can use URT's feature and improve their performance, further demonstrating the flexibility of our feature-agnostic approach.

\keypoint{Bayesian Few-Shot Meta-Learning} 
Relatively few methods in the literature take Bayesian approaches to few-shot learning. A few studies \cite{grant2018bayesMAML,yoon2018bayesian}  focus on understanding MAML \cite{finn2017model} as a hierarchical Bayesian model. 
Versa \cite{gordon2019metaPred} treats the weights of the final linear classifier layer as the quantity to infer given the support set during meta-test. It takes an amortized variational inference (VI) approach, training an inference neural network to predict the classifier parameters given the support set. However, unlike us, it then performs end-to-end representation learning, and is not fully Bayesian as it does not ultimately integrate the classifier parameters, as we achieve here. Neural Processes \cite{garnelo2018cnp} takes a Gaussian Process (GP) inspired approach to neural network design, but ultimately does not provide a clear Bayesian model. The recent DKT \cite{patacchiola2020bayesian} achieves true Bayesian meta-learning via GPs with end-to-end feature learning. However, despite performing feature learning, these Bayesian approaches have generally not provided SotA benchmark performance compared to the broader landscape of competitors at the time of their publication. A classic study \cite{heskes2000empirical} explored shallow learning-to-learn of linear regression by conjugacy. We also exploit conjugacy but for classifier learning, 
and demonstrate SotA results on heavily benchmarked tasks for the first time with Bayesian meta-learning. 

\keypoint{Classifier Layer Design} The vast majority of few-shot studies use either linear \cite{mangla2020charting,gordon2019metaPred,dvornik2020selecting}, cosine similarity \cite{qiao2017few}, or nearest centroid classifiers \cite{wang2019simpleshot,snell2017prototypical} under some distance metric. We differ in: (i) using a quadratic classifier, and (ii) taking a ``generative'' approach to fitting the model \cite{hastie2009elements}. While a quadratic classifier potentially provides a stronger fit than a linear classifier, its larger number of parameters will overfit catastrophically in a few-shot/high-dimension regime. This is why few studies have applied them, with the exception of \cite{bateni2020improved} who had to carefully hand-craft regularisers for them. Our key insight is to use conjugacy to enable the quadratic classifier prior to be efficiently meta-learned, thus gaining improved fitting strength, while avoiding overfitting.

\section{Probabilistic Meta-Learning}
\label{sec:probabilistic-meta-learning}
One can formalise a conventional classification problem as consisting of an input space $\mathcal{X}$, an output space $\mathcal{Y}$, and a distribution $p$ over $\mathcal{X} \times \mathcal{Y}$ that defines the task to be solved. Few-shot recognition is the problem of training a classifier to distinguish between $C$ different classes in a sparse data regime, where only $K$ labelled training instances are available for each class. Meta-learning aims to distill relevant knowledge from multiple related few-shot learning problems into a set of shared parameters that boost the learning of subsequent novel few-shot tasks. The simplest way to extend the standard formalisation of classification problems to a meta-learning context is to instead consider the set, $\mathcal{P}$ of all distributions over $\mathcal{X} \times \mathcal{Y}$, each of which represents a possible classification task. One can then assume the existence of a distribution, $Q$ over $\mathcal{P}$~\cite{baxter2000model}.

From a probabilistic perspective, the parameters inferred by the meta-learner that are shared across tasks, which we denote by $\phi$, can be seen as specifying or inducing a prior distribution over the task-specific parameters for each few-shot problem. As such, meta-learning can be thought of as learning a procedure to induce a prior over models for future tasks by meta-training on a collection of related tasks. Representing task-specific parameters for task $t$ by $\theta_t$, the few-shot training (aka support) and testing (aka query) sets as $D_S^t$ and $D^t_Q$, a Bayesian few-shot learner should use the learned prior to determine the posterior distribution over model parameters,
\begin{equation}
\label{eq:parameter-posterior}
    p(\theta_t | D_S^t, \phi) = \frac{p(D_S^t | \theta_t) p(\theta_t | \phi)}{\int p(D_S^t | \theta_t) p(\theta_t | \phi) \textup{d}\theta_t}.
\end{equation}
Once the distribution obtained, one can model query samples, $(\vec x_i^t, y_i^t)\in D_Q^t$, using the posterior predictive distribution,
\begin{equation}
\label{eq:posterior-predictive}
    p(D_Q^t | D_S^t, \phi) = \prod_{i=1}^{|D_Q^t|} \int p(\vec x_i^t, y_i^t | \theta_t) p(\theta_t | D_S^t, \phi) \textup{d}\theta_t.
\end{equation}
A natural measure for the goodness of fit for $\phi$ is the expected log likelihood of the few-shot models that make use of the shared prior,
\begin{equation}
    \underset{D_S, D_Q \sim q, q \sim Q}{\mathbb{E}} \lbrack L(\phi | D_S, D_Q) \rbrack,
\end{equation}
where
\begin{equation}
    L(\phi | D_S, D_Q) = \sum_{i=1}^{|D_Q|} \textup{log} \, p(\vec x_i, y_i | D_S, \phi).
\end{equation}
The process of meta-learning the prior parameters can then be formalised as an risk minimisation problem,
\begin{equation}
\label{eq:meta-objective}
    \phi^\ast = \argmin_{\phi} \underset{D_S, D_Q \sim q, q \sim Q}{\mathbb{E}} \lbrack -L(\phi | D_S, D_Q) \rbrack.
\end{equation}

\keypoint{Discussion} A prior probabilistic meta-learner \cite{gordon2019metaPred} focused on the term $p(\theta_t | D_S^t, \phi)$, taking an amortized variational inference perspective that treats $\phi$ as the parameters of a neural network that predicts a distribution over the parameters $\theta_t$ of a linear classifier given support set $D_S^t$. In contrast, our framework will use a QDA rather than linear classifier, and then exploit conjugacy to efficiently compute a distribution over the QDA mean and covariance parameters $\theta_t$ given the support set. This is both efficient and probabilistically cleaner, as our model contains a proper prior, while \cite{gordon2019metaPred}  does not.

The integrals in Eqs.~\ref{eq:parameter-posterior} and \ref{eq:posterior-predictive} are key to Bayesian meta-learning, but can be computationally intractable and \cite{gordon2019metaPred} relies on sampling. Our conjugate setup allows the integrals to be computed exactly in closed form, without relying on sampling. 

\section{Meta-Quadratic Discriminant Analysis}
Our MetaQDA provides a meta-learning generalization of the classic QDA classifier \cite{hastie2009elements}. QDA works by constructing a multivariate Gaussian distribution $\theta$ corresponding to each class by maximum likelihood. At test time, predictions are made by computing the likelihood of the query instance under each of these distributions, and using Bayes theorem to obtain the posterior $p(y|x,\theta)$. 
Rather than using maximum likelihood fitting for meta-testing, we introduce a Bayesian version of QDA that will enable us to exploit a meta-learned prior over the parameters of the multivariate Gaussian distributions.
Two Bayesian strategies for inference using such a prior are explored: 1) using the maximum a posterior (MAP) estimate of the Gaussian parameters; and 2) the fully Bayesian approach that propagates the parameter uncertainty through to the class predictions. The first of these is conceptually simpler, while the second allows for better handling of uncertainty due to the fully Bayesian nature of the parameter inference. For both cases we make use of Normal-Inverse-Wishart priors \cite{gelman2003bda}, as their conjugacy with multivariate Gaussians leads to an efficient implementation strategy. 

\subsection{MAP-Based QDA}\label{sec:mqdaMAP}
\label{sec:method}
We begin by describing a MAP variant of QDA. In conventional QDA the likelihood of an instance, $\vec x \in \mathbb{R}^d$, belonging to class $j \in \mathbb{N}_C$ is given by $\Gauss(\vec x | \vec \mu_j, \Sigma_j)$
and the parameters are found via maximum likelihood estimation (MLE) on the subset of the support set associated with class $j$,
\begin{equation}
    \vec \mu_j, \Sigma_j = \argmax_{\vec \mu, \Sigma} \prod_{i=1}^K \Gauss(\vec x_{j,i} | \vec \mu, \Sigma).
\end{equation}
This optimisation problem has a convenient closed form solution: the sample mean and covariance of the relevant subset of the support set. In order to incorporate prior knowledge learned from related few-shot learning tasks, we define a Normal-inverse-Wishart (NIW) prior \cite{murphy2012machine} over the parameters and therefore obtain a posterior for the parameters,
\begin{align}
\begin{split}
    &p(\vec \mu_j, \Sigma_j | \vec x, \vec m, \kappa, S, \nu) \\
    &= \frac{\prod_{i=1}^K\Gauss(\vec x_{j,i} | \vec \mu_j, \Sigma_j) \NIW(\vec \mu_j, \Sigma_j | \vec m, \kappa, S, \nu)}{\int \int \prod_{i=1}^K\Gauss(\vec x_{j,i} | \vec \mu, \Sigma) \NIW(\vec \mu^\prime, \Sigma^\prime | \vec m, \kappa, S, \nu) d\vec \mu^\prime d\Sigma^\prime}.
\end{split}
\end{align}

\keypoint{Training}
This enables us to take advantage of prior knowledge learned from related tasks when inferring the model parameters by MAP inference,
\begin{equation}
    \vec \mu_j, \Sigma_j = \argmax_{\vec \mu, \Sigma} \prod_{i=1}^K p(\vec \mu_j, \Sigma_j | \vec x_{j,i}, \vec m, \kappa, S, \nu).
\end{equation}

Because NIW is the conjugate prior of multivariate Gaussians, we know that the posterior distribution over the parameters takes the form of
\begin{equation}
    p(\vec \mu_j, \Sigma_j | \vec x, \vec m, \kappa, S, \nu) = \NIW(\vec \mu_j, \Sigma_j | \vec m_j, \kappa_j, S_j, \nu_j),
\end{equation}
where
\begin{align}
\begin{split}
\label{eq:update-rule}
    \vec m_j &= \frac{\vec m + K \hat{\vec \mu}_j}{\kappa + K}, \quad \kappa_j = \kappa + K,\quad \nu_j = \nu + K,  \\
    S_j &= S + \sum_{i=1}^K (\vec x_{j,i} - \hat{\vec \mu}_j)(\vec x_{j,i} - \hat{\vec \mu}_j)^T + \\
    &\quad \frac{\kappa K}{\kappa + K} (\hat{\vec \mu}_j - \vec m)(\hat{\vec \mu}_j - \vec m)^T,   
\end{split}
\end{align}
and we have used $\hat{\vec \mu}_j = \frac{1}{k}\sum_{i=1}^K \vec x_{j,i}$. The posterior is maximised at the mode, which occurs at
\begin{equation}
    \label{eq:map-parameters}
    \vec \mu_j = \vec m_j, \quad \Sigma_j = \frac{1}{\nu_j + d + 1} S_j.
\end{equation}

\keypoint{Testing}
After computing point estimates of the parameters, one can make predictions on instances from the query set according to the usual QDA model,
\begin{equation}
    p(y=j | \vec x, \vec m, \kappa, S, \nu) = \frac{\Gauss(\vec x | \vec \mu_j, \Sigma_j) p(y=j)}{\sum_{i=1}^C \Gauss(\vec x | \vec \mu_i, \Sigma_i) p(y=i)}.\label{eq:qda-test}
\end{equation}
Note the prior over the classes $p(y)$ can be dropped in the standard few-shot benchmarks that assume a uniform distribution over classes.

\subsection{Fully Bayesian QDA}\label{sec:mqdaFB}
Computing point estimates of the parameters throws away potentially useful uncertainty information that can help to better calibrate the predictions of the model. Instead, we can marginalise the parameters out when making a prediction,
\begin{align}
\begin{split}
    &p(y=j | \vec x) \\
    &= \frac{\int \int \Gauss(\vec x | \mu_j, \Sigma_j) \NIW(\vec \mu_j, \Sigma_j | \vec m_j, \kappa_j, S_j, \nu_j) d\vec \mu_j d \Sigma_j}{\sum_{i=1}^C \int \int \Gauss(\vec x | \mu_i, \Sigma_i) \NIW(\vec \mu_i, \Sigma_i | \vec m_j, \kappa_j, S_j, \nu_j) d\vec \mu_i d \Sigma_i}.
\end{split}
\end{align}
Each of the double integrals has the form of a multivariate $t$-distribution~\cite{murphy2012machine}, yielding
\begin{align}
\begin{split}
    &p(y=j | \vec x, \vec m, \kappa, S, \nu) \\
    &= \frac{\T \Big (\vec x | \vec m_j, \frac{\kappa_j + 1}{\kappa_j(\nu_j - d + 1)}S_j, \nu_j - d + 1 \Big )}{\sum_{i=1}^C \T \Big (\vec x | \vec m_i, \frac{\kappa_i + 1}{\kappa_i(\nu_i - d + 1)}S_i, \nu_i - d + 1 \Big )}.\label{eq:studentt-test}
\end{split}
\end{align}

\subsection{Meta-Learning the Prior}
Letting $\phi = (\vec m, \kappa, S, \nu)$, our objective is to minimise the negative expected log likelihood of models constructed with the shared prior on the parameters, as given in Eq.~\ref{eq:meta-objective}. For  MAP-based QDA, the log likelihood function is given by
\begin{equation}
\label{eq:map-likelihood}
    L(\phi | D_S, D_Q) = \sum_{j=1}^C \sum_{i=1}^K \textup{log} \, \Gauss(\vec x_{j,i} | \vec \mu_j, \Sigma_j),
\end{equation}
where $\vec \mu_j$ and $\Sigma_j$ are the point estimates computed via the closed-form solution to the MAP inference problem given in Equation~\ref{eq:map-parameters}. When using the fully Bayesian variant of QDA, we have the following log likelihood function:
\begin{align}
\begin{split}
\label{eq:full-likelihood}
    &L(\phi | D_S, D_Q) \\
    &= \sum_{j=1}^C \sum_{i=1}^K \textup{log} \, \T \Big (\vec x_{j,i} | \vec m_j, \frac{\kappa_j + 1}{\kappa_j(\nu_j - d + 1)}S_j, \nu_j - d + 1 \Big ).
\end{split}
\end{align}

\keypoint{Meta-Training}
We approximate the optimization in Equation~\ref{eq:meta-objective} by performing empirical risk minimisation on a training dataset using episodic training. In particular, we choose $\mathcal{P}$ to be the set of uniform distributions over all possible $C$-way classification problems, $Q$ as the uniform distribution over $\mathcal{P}$, and the process of sampling from each $q \in \mathcal{P}$ results in balanced datasets containing $K$ instances from each of the $C$ classes. Episodic training then consists of sampling a few-shot learning problem, building a Bayesian QDA classifier using the support set, computing the negative log likelihood on the query set, and finally updating $\phi$ using stochastic gradient descent. Crucially, the use of conjugate priors means that no iterative optimisation procedure must be carried out when constructing the classifier in each episode. Instead, we are able to backpropagate through the conjugacy update rules and directly modify the prior parameters with stochastic gradient descent. The overall learning procedure is given in Algorithm~\ref{alg:main}.

Some of the prior parameters must be constrained in order to learn a valid NIW distribution. In particular, $S$ must be positive definite, $\kappa$ must be positive, and $\nu$ must be strictly greater than $d-1$. The constraints can be enforced for $\kappa$ and $\nu$ by clipping any values that are outside the valid range back to the minimum allowable value. We parameterise the scale matrix in terms of its Cholesky factors,
\begin{equation}\label{eq:decomp}
    S = LL^T,
\end{equation}
where $L$ is a lower triangular matrix. During optimisation we ensure $L$ remains lower triangular by setting all elements above the diagonal to zero after each weight update.

\begin{algorithm}[t]
\caption{Pseudocode for epsiodic meta-learning of hyper-parameters in MetaQDA.}\label{alg:main}
\textbf{Require:} Distribution over tasks $Q$, number of iterations $T$, learning rate $\alpha$ \\
\textbf{Result:} prior parameters $\phi_T$ \\
\textbf{Init:} $\phi_0 = \{ \vec m = \vec{0}, S = \bold{I}, \kappa=1, \nu=d\}$ \\
\For{$t =1 \textup{ to } T$}{
    Sample task, $q_t \sim Q$ \;
    Sample support and query set, $D_S^t, D_Q^t \sim q_t$ \;
\textbf{Build Bayesian QDA Model} \\
    \text{If MAP:} $\theta_t \gets \{(\vec \mu_j, \Sigma_j)\}_{j=1}^C$ \tcp*{Eq~\ref{eq:map-parameters}}
    \text{If Fully Bayes:} $\theta_t \gets \{(\vec m_j, \kappa_j, S_j, \nu_j)\}_{j=1}^C$ \tcp*{Eq~\ref{eq:update-rule}}
\textbf{Update Prior} \\
    $\phi_{t} \gets \phi_{t-1} - \alpha \nabla_{\phi} L(\phi_{t-1} | D_S^t, D_Q^t)$ \tcp*{Eq~\ref{eq:map-likelihood} or \ref{eq:full-likelihood}}
}
\end{algorithm}

\section{Experiments}

We measure the efficacy of our model in standard, cross-domain and multi-domain few-shot learning and few-shot class-incremental problem settings. 
MetaQDA is a shallow classifier-layer meta-learner that is agnostic to the choice of fixed extracted features. 
Unless otherwise stated, we report results for the FB-based variant of MetaQDA.
During meta-training, we learn the priors $\phi = (\vec m, \kappa, S, \nu)$ over episodes drawn from the training set, keeping the feature extractor fixed. We use the meta-validation datasets for model selection and hyperparameter tuning. 
During meta-testing, the support set is used to obtain the parameter posterior, and then a QDA classifier is established according to either Eq~\ref{eq:qda-test} or Eq~\ref{eq:studentt-test}. All algorithms are evaluated on $C$-way $k$-shot learning \cite{snell2017prototypical}, with a batch of 15 query images per class in a testing episode. All accuracies are calculated by averaging over 600 randomly generated testing tasks with 95\% confidence interval.

\subsection{Standard Few-Shot Learning}

\keypoint{Datasets} 
\textbf{\miniIN} \cite{ravi2017optimization} is split into 64/16/20 for meta-train/val/test, respectively, containing 100 classes and 600 examples per class, drawn from ILSVRC-12 \cite{russakovsky2015imagenet}. Images are resized to 84$\times$84 \cite{he2016deep}.
\textbf{\tierIN} is a more challenging benchmark \cite{ren2018meta} consisting of 608 classes (779,165 images) and 391/97/160 classes for meta-train/val/test folds, respectively. Images are resized to 84$\times$84.
\textbf{CIFAR-FS} \cite{bertinetto2019R2D2} was created by randomly sampling from CIFAR-100 \cite{krizhevsky2009learning} by using the same criteria as \miniIN{} (100 classes with 600 images per class, split into folds of 64/16/20 for meta-train/val/test). Image are resized to 32$\times$32.

\keypoint{Feature Extractors} 
\textbf{Conv-4} (64-64-64-64) as in \cite{snell2017prototypical}. 
See Appendix for details. 
\textbf{ResNet-18} is standard 18-layer 8-block architecture with pre-trained weights in \cite{wang2019simpleshot}. \textbf{WRN-28-10} is standard architecture with 28 convolutional layers and width factor 10, and the pre-trained weights from \cite{mangla2020charting}.

\keypoint{Competitors}
We group competitors into two categories: 1) direct competitors that also make use of ``off-the-shelf'' fixed pre-trained networks and only update the classifier to learn novel classes; and 2) non-direct competitors that specifically meta-learn a feature optimised for few-shot learning and/or update features during meta-testing. 
Baseline++ \cite{chen2019closerfewshot} fixes the feature encoder and only tunes the (cosine similarity) classifier during the meta-test stage. SimpleShot \cite{wang2019simpleshot} uses an NCC classifier with different feature encoders and studies different feature normalizations. We use their best reported variant, CL2N. 
S2M2 \cite{mangla2020charting} uses a linear classifier after self-supervised and/or regularized classifier pre-training. SUR \cite{dvornik2020selecting} also uses pre-trained feature extractors, but focuses on weighting multiple features extracted from different backbones or multiple layers of the same backbone. We compare their reported results of a single ResNet backbone trained for multi-class classification as per ours, but they have the advantage of fusing features extracted from multiple layers. 
Unravelling \cite{goldblum2020unraveling} proposes some new regularizers for vanilla backbone training that improve feature quality for few-shot learning without meta-learning.

\setlength{\tabcolsep}{4.8pt}
\begin{table}[h]
\centering
\footnotesize
\resizebox{1.0\columnwidth}{!}{%
\begin{tabular}{@{} llcc @{}}
\toprule
\bf Model  & \bf Backbone & 1-shot & 5-shot\\
\midrule 

\textbf{\textsc{MetaLSTM}} \cite{ravi2017optimization} &Conv-4  &43.44 $\pm$ 0.77\% & 60.60 $\pm$ 0.71\% \\ 
\textbf{\textsc{MAML}}$^O$ \cite{finn2017model} & Conv-4 &  48.70 $\pm$ 1.84\% & 63.11 $\pm$ 0.92\% \\ 
\textbf{\textsc{ProtoNet}} \cite{snell2017prototypical} & Conv-4 &49.42 $ \pm $ 0.78\% &68.20 $\pm$ 0.66\%  \\
\textbf{\textsc{GNN}} \cite{garcia2017few} & Conv-4 & 50.33 $\pm$ 0.36\% & 66.41 $\pm$ 0.63\% \\ 
\textbf{\textsc{MetaSSL}}\cite{ren2018meta}&  Conv-4 &50.41 $\pm$ 0.31\% & 64.39 $\pm$ 0.24\% \\ 
\textbf{\textsc{RelationNet}} \cite{sung2018learning}& Conv-4 & 50.44 $\pm$ 0.82\% &65.32 $\pm$ 0.70\% \\ 
\textbf{\textsc{MetaSGD}}$^O$ \cite{li2017meta}  & Conv-4 & 50.47 $\pm$ 1.87\% & 64.03 $\pm$ 0.94\% \\ 
\textbf{\textsc{CAVIA}} \cite{zintgraf2018cavia} & Conv-4 & 51.82 $\pm$ 0.65\% & 65.85 $\pm$ 0.55\%\\
\textbf{\textsc{TPN}} \cite{liu2018transductive} & Conv-4 & 52.78 $\pm$ 0.27\% & 66.59 $\pm$ 0.28\% \\ 
\textbf{\textsc{R2D2}} \cite{bertinetto2019R2D2} & Conv-4$^*$ & {51.90 $\pm$ 0.20\%} & {68.70 $\pm$ 0.20\%} \\
\textbf{\textsc{RelationNet2}}\cite{xueting2020dcn} & Conv-4 & {53.48 $\pm$ 0.78\%} & {67.63 $\pm$ 0.59\%} \\
\textbf{\textsc{GCR}} \cite{li2019GCR} & Conv-4 &53.21 $ \pm $ 0.40\% & 72.34 $\pm$ 0.32\%  \\
\textbf{\textsc{VERSA}} \cite{gordon2019metaPred} & Conv-4 &53.40 $ \pm $ 1.82\% & 67.37 $\pm$ 0.86\%  \\
\textbf{\textsc{DynamicFSL}}$^\dagger$ \cite{gidaris2018dynamic} & Conv-4 & 56.20 $\pm$ 0.86\% & 72.81 $\pm$ 0.62\%\\

\rowcolor{Gray} \textbf{\textsc{Baseline++}} \cite{chen2019closerfewshot} & Conv-4 &  48.24 $\pm$ 0.75\% & 66.43 $\pm$ 0.63\% \\ 
\rowcolor{Gray} \textbf{\textsc{SimpleShot}}\cite{wang2019simpleshot} & Conv-4 & {49.69 $\pm$ 0.19\%} & {66.92 $\pm$ 0.17\%} \\
\rowcolor{Gray} \textbf{\textsc{MetaQDA}} & Conv-4 & \textbf{56.41 $\pm$ 0.80\%} & \textbf{72.64 $\pm$ 0.62\%} \\ 

\midrule

\textbf{\textsc{SNAIL}} \cite{santoro2017simple} & ResNet-12 & 55.71 $\pm$ 0.99\% & 68.88 $\pm$ 0.92\% \\ 
\textbf{\textsc{Dynamic FSL}} \cite{gidaris2018dynamic} & ResNet-12& 55.45 $\pm$ 0.89\% & 70.13 $\pm$ 0.68\% \\ 
\textbf{\textsc{TADAM}} \cite{oreshkin2018tadam} & ResNet-12 & 58.50 $\pm$ 0.30\% & {76.70 $\pm$ 0.30\%} \\
\textbf{\textsc{CAML}} \cite{jiang2019CAML} & ResNet-12 &  59.23 $\pm$ 0.99\% & 72.35 $\pm$ 0.18\% \\
\textbf{\textsc{AM3}} \cite{xing2019am3} & ResNet-12 & 65.21 $\pm$ 0.49\% & 75.20 $\pm$ 0.36\% \\
\textbf{\textsc{MTL}} \cite{sun2019meta} & ResNet-12$^*$ & 61.20 $\pm$ 1.80\% & 75.50 $\pm$ 0.80\% \\
\textbf{\textsc{Tap Net}} \cite{yoon2019tapnet} & ResNet-12 & 61.65 $\pm$ 0.15\% & 76.36 $\pm$ 0.10\% \\
\textbf{\textsc{RelationNet2}}\cite{xueting2020dcn} & ResNet-12 & 63.92 $\pm$ 0.98\% & 77.15 $\pm$ 0.59\% \\
\textbf{\textsc{R2D2}}\cite{bertinetto2019R2D2} & ResNet-12 & 59.38 $\pm$ 0.31\% & 78.15 $\pm$ 0.24\% \\
\textbf{\textsc{MetaOpt}}$^{O}$ \cite{lee2019meta} & ResNet-12$^*$ & 64.09 $\pm$ 0.62\% & 80.00 $\pm$ 0.45\% \\

\textbf{\textsc{RelationNet}} \cite{chen2019closerfewshot} & ResNet-18 &  52.48 $\pm$ 0.86\% & 69.83 $\pm$ 0.68\% \\
\textbf{\textsc{ProtoNet}} \cite{chen2019closerfewshot} & ResNet-18 &  54.16 $\pm$ 0.82\% & 73.68 $\pm$ 0.65\% \\
\textbf{\textsc{DCEM}} \cite{dvornik2019dcem} & ResNet-18 &  58.71 $\pm$ 0.62\% & 77.28 $\pm$ 0.46\% \\
\textbf{\textsc{AFHN}} \cite{li2020afhn} & ResNet-18 &  62.38 $\pm$ 0.72\% & 78.16 $\pm$ 0.56\% \\

\rowcolor{Gray} \textbf{\textsc{SUR}}\cite{dvornik2020selecting} & ResNet-12 & 60.79 $\pm$ 0.62\% & 79.25 $\pm$ 0.41\% \\
\rowcolor{Gray} \textbf{\textsc{Unravelling}}\cite{goldblum2020unraveling} & ResNet-12$^*$
& {59.37 $\pm$ 0.32\%} & {77.05 $\pm$ 0.25\%} \\
\rowcolor{Gray} \textbf{\textsc{Baseline++}} \cite{chen2019closerfewshot} & ResNet-18 &  51.87 $\pm$ 0.77\% & 75.68 $\pm$ 0.63\% \\ 
\rowcolor{Gray} \textbf{\textsc{SimpleShot}}\cite{wang2019simpleshot} & ResNet-18 & {62.85 $\pm$ 0.20\%} & {80.02 $\pm$ 0.14\%} \\
\rowcolor{Gray} \textbf{\textsc{S2M2}} \cite{mangla2020charting} & ResNet-18 &  64.06 $\pm$ 0.18\% & 80.58 $\pm$ 0.12\% \\
\rowcolor{Gray} \textbf{\textsc{MetaQDA}} & ResNet-18 & \textbf{ 65.12 $\pm$ 0.66\%} & \textbf{80.98 $\pm$ 0.75\%} \\ 

\midrule 

\textbf{\textsc{LEO}}$^{O}$ \cite{rusu2019leo} & WRN & 61.78 $\pm$ 0.05\% & 77.59 $\pm$ 0.12\% \\
\rowcolor{Gray} \textbf{\textsc{PPA}} \cite{qiao2017few} & WRN & 59.60 $\pm$ 0.41\% & 73.74 $\pm$ 0.19\% \\ 
\rowcolor{Gray} \textbf{\textsc{SimpleShot}}\cite{wang2019simpleshot} & WRN & {63.50 $\pm$ 0.20\%} & {80.33 $\pm$ 0.14\%} \\
\rowcolor{Gray} \textbf{\textsc{S2M2}} \cite{mangla2020charting} & WRN & 64.93 $\pm$ 0.18\% & 83.18 $\pm$ 0.22\% \\
\rowcolor{Gray} \textbf{\textsc{MetaQDA}} & WRN & \textbf{67.83 $\pm$ 0.64\%} & \textbf{84.28 $\pm$ 0.69\%} \\ 

\bottomrule
\end{tabular}
}
\caption{\small \small
\textbf{Few-shot classification results on \miniIN{}.} 
$^\dagger$: two-step optimization with  attention. 
$^O$: requires gradient-based optimisation at meta-test time. $^*$: Use a wider CNN than standard and higher dimensional embedding. 
Grey: Fixed feature methods.}
\label{tab:fsl_mini}
\end{table}

\keypoint{Results}
Table~\ref{tab:fsl_mini}-\ref{tab:fsl_cifarfs} summarize the results on \miniIN{}, \tierIN{} and CIFAR-FS. MetaQDA performs better than all the previous methods that rely on off-the-shelf feature extractors, and also the majority of methods that meta-learn representations specialised for few-shot problems. We do not make efforts to carefully fine-tune the hyperparameters, but focus on showing that our model has robust advantages in different few-shot learning benchmarks with various backbones. A key benefit of fixed feature approaches (grey) is small \textbf{compute cost}, e.g., under 1-hour training. In contrast, SotA end-to-end competitors (white) such as \cite{lee2019meta,gidaris2018dynamic,xueting2020dcn} require over 10 hours.



\setlength{\tabcolsep}{4.8pt}
\begin{table}[t]
\centering
\footnotesize
\resizebox{1.0\columnwidth}{!}{
\begin{tabular}{@{} llcc @{}}
\toprule
\bf Model  & \bf Backbone & 1-shot & 5-shot \\
\midrule 

\textbf{\textsc{MAML}} \cite{liu2018transductive} & Conv-4&  51.67 $\pm$ 1.81\% & 70.30 $\pm$ 1.75\% \\ 
\textbf{\textsc{MetaSSL}}$^\dagger$ \cite{ren2018meta}&Conv-4& 52.39 $\pm$ 0.44\% & 70.25 $\pm$ 0.31\% \\
\textbf{\textsc{RelationNet}} \cite{liu2018transductive} &Conv-4& 54.48 $\pm$ 0.48\% &71.31 $\pm$ 0.78\% \\ 
\textbf{\textsc{TPN}}$^\dagger$ \cite{liu2018transductive}&Conv-4& 59.91 $\pm$ 0.94\% & 73.30 $\pm$ 0.75\% \\
\textbf{\textsc{RelationNet2}}\cite{xueting2020dcn} & Conv-4 & 60.58 $\pm$ 0.72\% & 72.42 $\pm$ 0.69\% \\
\textbf{\textsc{ProtoNet}} \cite{liu2018transductive} &Conv-4 &53.31 $\pm$ 0.89\%  &72.69 $\pm$ 0.74\% \\ 
\rowcolor{Gray} \textbf{\textsc{SimpleShot}}\cite{wang2019simpleshot} & Conv-4 & {51.02 $\pm$ 0.20\%} & {68.98 $\pm$ 0.18\%} \\
\rowcolor{Gray} \textbf{\textsc{MetaQDA}} & Conv-4 & \textbf{58.11 $\pm$ 0.48\%} & \textbf{74.28 $\pm$ 0.73\%} \\ 

\midrule 

\textbf{\textsc{TapNet}} \cite{yoon2019tapnet} & ResNet-12& 63.08 $\pm$ 0.15\% & 80.26 $\pm$ 0.12\% \\
\textbf{\textsc{RelationNet2}} \cite{xueting2020dcn} & ResNet-12 & 68.58 $\pm$ 0.63\%  & 80.65 $\pm$ 0.91\%\\
\textbf{\textsc{MetaOptNet}}$^O$ \cite{lee2019meta} & ResNet-12$^*$ & 65.81 $\pm$ 0.74\% & 81.75 $\pm$ 0.53\%\\
\rowcolor{Gray} \textbf{\textsc{SimpleShot}} \cite{wang2019simpleshot} & ResNet-18 & {69.09 $\pm$ 0.22\%} & {84.58 $\pm$ 0.16\%} \\
\rowcolor{Gray} \textbf{\textsc{MetaQDA}} & ResNet-18 & \textbf{69.97 $\pm$ 0.52\%} & \textbf{85.51 $\pm$ 0.58\%} \\ 

\midrule 
\textbf{\textsc{LEO}} \cite{rusu2019leo} & WRN & 66.33 $\pm$ 0.05\% & 81.44 $\pm$ 0.09\% \\
\rowcolor{Gray} \textbf{\textsc{SimpleShot}} \cite{wang2019simpleshot} & WRN & {69.75 $\pm$ 0.20\%} & {85.31 $\pm$ 0.15\%} \\
\rowcolor{Gray} \textbf{\textsc{S2M2}} \cite{mangla2020charting} & WRN & 73.71 $\pm$ 0.22\% & 88.59 $\pm$ 0.14\% \\
\rowcolor{Gray} \textbf{\textsc{MetaQDA}} & WRN & \textbf{74.33 $\pm$ 0.65\%} & \textbf{89.56 $\pm$ 0.79\%} \\ 

\bottomrule
\end{tabular}%
}
\caption{\small \small
\textbf{Few-shot classification results on \tierIN{}.} 
$^\dagger$: Make use of additional unlabeled data for semi-supervised learning or transductive inference.  
Gray: Use fixed pre-trained backbones.
}
\label{tab:fsl_tiered}
\end{table}

\setlength{\tabcolsep}{4.8pt}
\begin{table}[t]
\centering
\footnotesize
\resizebox{1.0\columnwidth}{!}{
\begin{tabular}{@{} llcc @{}}
\toprule
\bf Model  & \bf Backbone & 1-shot & 5-shot \\
\midrule 

\textbf{\textsc{MAML}} \cite{mangla2020charting} & Conv-4 &  58.90 $\pm$ 1.90\%  & 71.50 $\pm$ 1.00\% \\ 
\textbf{\textsc{RelationNet}} \cite{mangla2020charting} & Conv-4 & 55.50 $\pm$ 1.00\% & 69.30 $\pm$ 0.80\% \\
\textbf{\textsc{ProtoNet}} \cite{mangla2020charting} & Conv-4 & 55.50 $\pm$ 0.70\% & 72.02 $\pm$ 0.60\% \\
\textbf{\textsc{R2D2}} \cite{bertinetto2019R2D2} & Conv-4 & 62.30 $\pm$ 0.20\% & 77.40 $\pm$ 0.10\% \\ 
\rowcolor{Gray} \textbf{\textsc{SimpleShot}}$^+$ \cite{wang2019simpleshot} & Conv-4 & { 59.35 $\pm$ 0.89\%} & { 74.76 $\pm$ 0.72\%} \\ 
\rowcolor{Gray} \textbf{\textsc{MetaQDA}} & Conv-4 & \textbf{60.52 $\pm$ 0.88\%} & \textbf{77.33 $\pm$ 0.73\%} \\ 
\midrule 
\textbf{\textsc{ProtoNet}} \cite{mangla2020charting} & ResNet-12 & 72.20 $\pm$ 0.70\% & 83.50 $\pm$ 0.50\% \\
\textbf{\textsc{MetaOpt}} \cite{lee2019meta} & ResNet-12$^*$ & 72.00 $\pm$ 0.70\% & 84.20 $\pm$ 0.50\% \\
\rowcolor{Gray} \textbf{\textsc{Unravelling}} \cite{goldblum2020unraveling} & ResNet-12$^*$ & 72.30 $\pm$ 0.40\% & 86.30 $\pm$ 0.20\% \\
\rowcolor{Gray} \textbf{\textsc{Baseline++}} \cite{chen2019closerfewshot,mangla2020charting} & ResNet-18 & 59.67 $\pm$ 0.90\% & 71.40 $\pm$ 0.69\% \\
\rowcolor{Gray} \textbf{\textsc{S2M2}} \cite{mangla2020charting} & ResNet-18 & 63.66 $\pm$ 0.17\% & 76.07 $\pm$ 0.19\% \\
\midrule

\textbf{\textsc{MetaOptNet}} \cite{lee2019meta} & WRN & 72.00 $\pm$ 0.70\% & 84.20 $\pm$ 0.50\% \\
\rowcolor{Gray} \textbf{\textsc{Baseline++}} \cite{mangla2020charting,chen2019closerfewshot} & WRN & 67.50 $\pm$ 0.64\% & 80.08 $\pm$ 0.32\% \\
\rowcolor{Gray} \textbf{\textsc{S2M2}} \cite{mangla2020charting} & WRN & 74.81 $\pm$ 0.19\% & 87.47 $\pm$ 0.13\% \\
\rowcolor{Gray} \textbf{\textsc{MetaQDA}} & WRN & \textbf{75.83 $\pm$ 0.88\%} & \textbf{88.79 $\pm$ 0.75\%} \\ 

\bottomrule
\end{tabular}%
}
\vspace{-1em}
\caption{\small \small
\textbf{Few-shot classification results on CIFAR-FS.}
$^+$  Our implementation. Gray: Use fixed pre-trained backbones.}
\label{tab:fsl_cifarfs}
\end{table}

\subsection{Cross-Domain Few-Shot Learning}

\keypoint{Dataset}
\textbf{CUB} \cite{hilliard2018few} contains 11,788 images across 200 fine-grained classes, split into folds of 100, 50, and 50 \textbf{Cars} \cite{krause2013cars,tseng2020cross} contains 196 classes randomly split into folds of 98, 49, and 49 classes for meta-train/val/test, respectively.

\keypoint{Competitors}
Better few-shot learning methods should degrade less when transferring to new domains \cite{chen2019closerfewshot,tseng2020cross}. 
We are specifically interested in comparing MetaQDA with other methods using off-the-shelf features. In particular, we consider \textit{Baseline++}~\cite{chen2019closerfewshot} and
\textit{S2M2}~\cite{mangla2020charting} who use linear classifiers, and the nearest centroid method of SimpleShot~\cite{wang2019simpleshot}. 

\keypoint{Results}
Table \ref{tab:cdfsl} demonstrates that MetaQDA exhibits good robustness to domain shift. Specifically, our method outperforms other approaches by at least $2\%-4\%$ across all dataset, support set size, and feature combinations.

\setlength{\tabcolsep}{4.8pt}
\begin{table}[t]
\centering
\footnotesize
\resizebox{1.0\columnwidth}{!}{%
\begin{tabular}{@{} cllcc @{}}
\\
\toprule
\\
&{\bf Model}  & {\bf Backbone} & 1-shot & 5-shot \\
\midrule \\
\parbox[t]{2mm}{\multirow{19}{*}{\rotatebox[origin=c]{90}{\miniIN{}$\to$CUB}}} & \textbf{\textsc{MAML}} \cite{patacchiola2020bayesian}& Conv-4 & 34.01 $\pm$ 1.25\% & - \\
&\textbf{\textsc{RelationNet}} \cite{patacchiola2020bayesian}& Conv-4 & 37.13 $\pm$ 0.20\% & - \\
&\textbf{\textsc{DKT}} \cite{patacchiola2020bayesian} & Conv-4 & 40.22 $\pm$ 0.54\% & - \\
&\textbf{\textsc{ProtoNet}} \cite{patacchiola2020bayesian} & Conv-4 & 33.27 $\pm$ 1.09\% & - \\

& \cellcolor{Gray} \textbf{\textsc{Baseline++}} \cite{patacchiola2020bayesian,chen2019closerfewshot} 
& \cellcolor{Gray}Conv-4 & \cellcolor{Gray}39.19 $\pm$ 0.12\% & \cellcolor{Gray}- \\
&\cellcolor{Gray}\textbf{\textsc{SimpleShot}}$^+$ \cite{wang2019simpleshot} &\cellcolor{Gray} Conv-4 &\cellcolor{Gray} 45.36 $\pm$ 0.75\% &\cellcolor{Gray} 61.44 $\pm$ 0.71\% \\
&\cellcolor{Gray}\textbf{\textsc{MetaQDA}} & \cellcolor{Gray}Conv-4 & \cellcolor{Gray}\textbf{47.25 $\pm$ 0.58\%} &\cellcolor{Gray} \textbf{64.40 $\pm$ 0.65\%} \\

\cmidrule{2-5} \\
&\textbf{\textsc{MAML}} \cite{chen2019closerfewshot} & ResNet-18 & - & 51.34 $\pm$ 0.72\% \\
&\textbf{\textsc{RelationNet}} \cite{chen2019closerfewshot} & ResNet-18 & - & 57.71 $\pm$ 0.73\% \\
&\textbf{\textsc{LRP (CAN)}} \cite{sun2020explanation} & ResNet-12 & 46.23 $\pm$ 0.42\% & 66.58 $\pm$ 0.39\% \\
&\textbf{\textsc{LRP (GNN)}} \cite{sun2020explanation} & ResNet-10 & 48.29 $\pm$ 0.51\% & 64.44 $\pm$ 0.48\% \\
&\textbf{\textsc{LFWT}} \cite{tseng2020cross} &ResNet-10 & 47.47 $\pm$ 0.75\% & 66.98 $\pm$ 0.68\% \\
&\textbf{\textsc{ProtoNet}} \cite{chen2019closerfewshot} & ResNet-18 & - & 62.02 $\pm$ 0.70\% \\

&\cellcolor{Gray}\textbf{\textsc{Baseline++}}
\cite{chen2019closerfewshot} & \cellcolor{Gray} ResNet-18 & \cellcolor{Gray}42.85 $\pm$ 0.69\%  & \cellcolor{Gray} 62.04 $\pm$ 0.76\% \\
&\cellcolor{Gray}\textbf{\textsc{SimpleShot}}$^+$ \cite{wang2019simpleshot} & \cellcolor{Gray}ResNet-18 &\cellcolor{Gray} 46.68 $\pm$ 0.49\% & \cellcolor{Gray}65.56 $\pm$ 0.70\% \\
&\cellcolor{Gray}\textbf{\textsc{MetaQDA}} & \cellcolor{Gray}ResNet-18 & \cellcolor{Gray}\textbf{48.88 $\pm$ 0.64\%} & \cellcolor{Gray} \textbf{68.59 $\pm$ 0.59\%} \\
\cmidrule{2-5}
&\cellcolor{Gray}\textbf{\textsc{S2M2}}  \cite{mangla2020charting} & \cellcolor{Gray} WRN &\cellcolor{Gray} 48.24 $\pm$ 0.84\% & \cellcolor{Gray}70.44 $\pm$ 0.75\% \\
&\cellcolor{Gray}\textbf{\textsc{SimpleShot}}$^+$ \cite{wang2019simpleshot} & \cellcolor{Gray}WRN & \cellcolor{Gray}49.65 $\pm$ 0.24\%  & \cellcolor{Gray}66.77 $\pm$ 0.19\% \\
&\cellcolor{Gray}\textbf{\textsc{MetaQDA}} &\cellcolor{Gray} WRN & \cellcolor{Gray} \textbf{53.75 $\pm$ 0.72\%} &\cellcolor{Gray} \textbf{71.84 $\pm$ 0.66\%}  \\

\bottomrule
\parbox[t]{2mm}{\multirow{10}{*}{\rotatebox[origin=c]{90}{\miniIN{}$\to$Cars}}} & \cellcolor{Gray}\textbf{\textsc{SimpleShot}}$^+$ \cite{wang2019simpleshot} & \cellcolor{Gray}Conv-4 & \cellcolor{Gray} 29.52 $\pm$ 0.56\% & \cellcolor{Gray}39.52 $\pm$ 0.66\%\\ 
&\cellcolor{Gray}\textbf{\textsc{MetaQDA}} & \cellcolor{Gray} Conv-4  &\cellcolor{Gray} \textbf{30.98 $\pm$ 0.66\%} &\cellcolor{Gray} \textbf{42.85 $\pm$ 0.68\%} \\ 
\cmidrule{2-5}
&\textbf{\textsc{LRP (CAN)}} \cite{sun2020explanation} & ResNet-12 &  32.66 $\pm$ 0.46\% & 43.86 $\pm$ 0.38\% \\
&\textbf{\textsc{LRP (GNN)}} \cite{sun2020explanation} & ResNet-10  & 32.78 $\pm$ 0.39\% & 46.20 $\pm$ 0.46\%  \\
&\textbf{\textsc{LFWT}} \cite{tseng2020cross} &ResNet-10  & 30.77 $\pm$ 0.47\% & 44.90 $\pm$ 0.64\%  \\
&\cellcolor{Gray}\textbf{\textsc{SimpleShot}}$^+$ \cite{wang2019simpleshot} & \cellcolor{Gray} ResNet-18 & \cellcolor{Gray} 34.72 $\pm$ 0.67\% & \cellcolor{Gray} \cellcolor{Gray} 47.26 $\pm$ 0.71\% \\
&\cellcolor{Gray}\textbf{\textsc{MetaQDA}} & \cellcolor{Gray}ResNet-18  & \cellcolor{Gray} \textbf{37.05 $\pm$ 0.65\%} & \cellcolor{Gray} \textbf{51.58 $\pm$ 0.52\%} \\ 

\cmidrule{2-5}
&\cellcolor{Gray}\textbf{\textsc{S2M2}}  \cite{mangla2020charting} & \cellcolor{Gray} WRN & \cellcolor{Gray} 31.52 $\pm$ 0.59\% & \cellcolor{Gray} 47.48 $\pm$ 0.68\% \ \\
&\cellcolor{Gray}\textbf{\textsc{SimpleShot}}$^+$ \cite{wang2019simpleshot} & \cellcolor{Gray}  WRN & \cellcolor{Gray} 33.68 $\pm$ 0.63\%  & \cellcolor{Gray} 46.67 $\pm$ 0.68\% \ \\
&\cellcolor{Gray}\textbf{\textsc{MetaQDA}} & \cellcolor{Gray} WRN &  \cellcolor{Gray} \textbf{36.21 $\pm$ 0.62\%} & \cellcolor{Gray} \textbf{50.83 $\pm$ 0.64\%} \\

\bottomrule \\
\end{tabular}
}

\caption{\small \small
\textbf{Cross domain few-shot classification results from \miniIN{} to CUB and Cars datasets.}
$^+$  Our implementation. Gray: Use fixed pre-trained backbones.
}
\vspace{-1.5em}
\label{tab:cdfsl}
\end{table}

\subsection{Multi-Domain Few-Shot Learning}

\keypoint{Dataset}
Meta-Dataset \cite{triantafillou2019meta} is a challenging large-scale benchmark spanning 10 image datasets. Following \cite{requeima2019cnaps, bateni2020improved}, we report results using the first 8 datasets for meta training (some classes are reserved for "in-domain" testing performance evaluation), and hold out entirely the remaining 2 (\textit{Traffic Signs} and \textit{MSCOCO}) plus an additional 3 datasets (\textit{MNIST} \cite{lecun2010mnist}, \textit{CIFAR10}, \textit{CIFAR100} \cite{krizhevsky2009learning}) for an unseen "out-of-domain" performance evaluation. Note that the meta-dataset protocol is random way and shot.

\keypoint{Competitors}  CNAP \cite{requeima2019cnaps} and SCNAP \cite{bateni2020improved} meta-learn an adaptive feature extractor whose parameters are modulated by an adaptation network that takes the current task’s dataset as input. SUR \cite{dvornik2020selecting}  performs feature selection among a suite of meta-train domain-specific features. The concurrent URT \cite{liu2020universal} meta-learns a transformer to dynamically meta-train dataset features before nearest-centroid classification with ProtoNet. We apply MetaQDA upon the fixed fused features learned by URT, replacing ProtoNet. 


\keypoint{Results} 
Table \ref{tab:metadataset} reports the average rank and accuracy of each model across all 13 datasets. We also break accuracy down among the `in-domain' and `out-of-domain' datasets (i.e., seen/unseen during meta-training). MetaQDA has the best average rank and overall accuracy. In particular it achieves strong out-of-domain performance, which is in line with our good cross-domain results above. Detailed results broken down by dataset are in the Appendix. 

\setlength{\tabcolsep}{4.8pt}
\begin{table}[h]
\centering
\footnotesize
\resizebox{1.0\columnwidth}{!}{%
\begin{tabular}{@{} lcccc @{}}
\toprule

\multirow{2}{*}{\bf Model} & \multicolumn{1}{c}{\bf Avg. Rank} & \multicolumn{3}{c}{\bf Avg. Accuracy}  \\
& {\bf overall} & {\bf overall} & {\bf in-domain} & {\bf out-of-domain} \\
\midrule 

\textbf{\textsc{CNAP}} \cite{requeima2019cnaps} & 4.5 & 65.9 $\pm$ 0.8\% & 69.6 $\pm$ 0.8\% & 59.8 $\pm$ 0.8\%\\
\textbf{\textsc{SCNAP}} \cite{bateni2020improved} & 2.9 & 72.2 $\pm$ 0.8\% & 73.8 $\pm$ 0.8\% & \bf{69.7 $\pm$ 0.8\%} \\
\rowcolor{Gray}\textbf{\textsc{SUR}} \cite{dvornik2020selecting} & 3.2 & 72.7 $\pm$ 0.9\%  & 75.6 $\pm$ 0.8\% & 68.1 $\pm$ 0.8\%\\ 
\rowcolor{Gray}\textbf{\textsc{URT+PN}} \cite{liu2020universal} & 2.4 & 73.7 $\pm$ 0.8\% & 77.2 $\pm$ 0.9\% & 68.1 $\pm$ 0.9\% \\ 
\rowcolor{Gray}\textbf{\textsc{URT+MQDA}} & 1.8 & \bf{74.3 $\pm$ 0.8\%} &\bf{77.7 $\pm$ 0.9\%}  & 68.8 $\pm$ 0.9\%  \\ 

\bottomrule
\end{tabular}%
}
\caption{\small \small
{\bf Few-shot classification results on Meta-Dataset.} Average accuracy and rank across episodes and datasets.}
\label{tab:metadataset}
\vspace{-1.5em}
\end{table}


\subsection{Few-Shot Class-Incremental Learning}
\keypoint{Problem Setup} 
Few-Shot Class-Incremental Learning (FSCIL) requires to \emph{incrementally} learn novel classes \cite{rebuffi2017icarl} from few labelled samples \cite{ren2019attentionAttractor,tao2020few} ideally without forgetting. Our fixed feature assumption provides both an advantage and a disadvantage in this regard. But our MetaQDA is naturally suited to incremental learning, and more than makes up for any disadvantage. Following \cite{tao2020few}, \miniIN{} is split into 60/40  base/novel classes, each with 500 training and 100 testing images. Each meta-test episode starts from a base classifier and proceeds in 8 learning sessions adding a 5-way-5-shot support set per session. After each session, models are evaluated on the full set of classes seen so far, leading to a 100-way generalized few-shot problem in the 9th session.

\keypoint{MetaQDA}
As per \cite{tao2020few}, we pre-train a ResNet18 backbone (see Appendix for details) and then meta-train MetaQDA on 60 base classes before performing incremental meta-testing. The MetaQDA prior is not updated during meta-testing.

\keypoint{Results}
Table~\ref{tab:FSCIL} reports the average results of 10 meta-test episodes with random 5-shot episodes. Clearly MetaQDA significantly outperforms both NCC and the previous SotA \cite{tao2020few}.

\setlength{\tabcolsep}{4.8pt}
\begin{table*}[h]
\centering
\footnotesize
\resizebox{1.0\textwidth}{!}{%
\begin{tabular}{@{} lccccccccc @{}}
\toprule
{\bf Model} & session 0 (60) & session 1 (65) & session 2 (70)	& session 3 (75)	& session 4 (80) & session 5 (85) & session 6 (90) & session 7 (95) & session 8 (100) \\
\midrule 
\textbf{\textsc{AL\_MML}} \cite{tao2020few}  & \textbf{61.31} & 50.09 & 45.17 & 41.16 & 37.48 & 35.52 & 32.19 & 29.46 & 24.42\\ 
\textbf{\textsc{NCC}} & 46.62 & 43.26& 40.87 & 39.04 & 37.50 & 35.96 & 34.13 & 33.19 & 32.26 \\ 
\textbf{\textsc{MetaQDA}} & 59.57 & \textbf{54.98} {\smaller (+4.89)} & \textbf{51.06} {\smaller (+5.89)} & \textbf{47.69} {\smaller (+6.53)} & \textbf{44.71} {\smaller (+7.23)} & \textbf{42.08} {\smaller (+6.56)} & \textbf{39.74} {\smaller (+7.55)} & \textbf{37.66} {\smaller (+8.20)} & \textbf{35.78} {\smaller (+11.36)}\\ 
\bottomrule
\end{tabular}}
\caption{\small \small
\textbf{Class-incremental few-shot learning with ResNet18 on \miniIN.} Start with 60-way base classifier and add 5-way/5-shot per session. At each session, the models are evaluated on the test sets of the full set of classes encountered so far. (\#): classifier-way at each session.
}
\label{tab:FSCIL}
\end{table*}

\subsection{Model Calibration}

In real world scenarios, where high-importance decisions are being made, the probability calibration of a machine learning model is critical \cite{guo2017calibration}. Any errors they make should be accompanied with associated low-confidence scores, e.g., so they can be checked by another process. 

\keypoint{Metrics} 
Following \cite{nixon2019measuring,guo2017calibration}, we compute Expected Calibration Error (ECE) with and without temperature scaling (TS). 
ECE assigns each prediction to a bin that indicates how confident the prediction is, which should reflect its probability of correctness. IE: $ECE=\sum_{b=1}^{B}\frac{n_b}{N}\left | \mathrm{acc}(b)-\mathrm{conf}(b) \right |$, where $n_b$ is the number of predictions in bin $b$, $N$ is the number of instances, and $\mathrm{acc}(b)$ and $\mathrm{conf}(b)$ are the accuracy and confidence of bin $b$. We use $B=20$. Temperature scaling uses validation episodes to calibrate a softmax temperature for best ECE. Please see \cite{nixon2019measuring,guo2017calibration} for full details.

\keypoint{Results}
Table \ref{tab:calibration} shows MetaQDA has superior uncertainty quantification compared to existing competitors. Vanilla QDA and SimpleShot are poorly calibrated, demonstrating the importance of our learned prior. The deeper WRN is also worse calibrated despite being more accurate, but MetaQDA ultimately compensates for this. Finally, we see that our fully-Bayesian (MetaQDA-FB, Sec~\ref{sec:mqdaFB}) variant outperforms our MAP (MetaQDA-MAP, Sec~\ref{sec:mqdaMAP}) variant.
\setlength{\tabcolsep}{4.8pt}
\begin{table}[tb]
\centering
\footnotesize
\resizebox{0.95\columnwidth}{!}{%
\begin{tabular}{@{} llcccc @{}}
\toprule
\multirow{2}{*}{\bf Model}  & \multirow{2}{*}{\bf Backbone} & \multicolumn{2}{c}{ECE+TS} & \multicolumn{2}{c}{ECE} \\
& & 1-shot & 5-shot & 1-shot & 5-shot \\

\midrule 
\textbf{\textsc{Lin.Classif.}} & Conv-4  & 3.56 & 2.88 & 8.54 &	7.48 \\
\textbf{\textsc{SimpleShot}} & Conv-4 & 3.82 & 3.35 & 33.45 & 45.81 \\
\textbf{\textsc{QDA}} & Conv-4 & 8.25 & 4.37 & 43.54 & 26.78 \\ 
\textbf{\textsc{MQDA-MAP}} & Conv-4 & 2.75 & 0.89 & 8.03 & 5.27\\
\textbf{\textsc{MQDA-FB}}& Conv-4 & \bf{2.33} & \bf{0.45} & \bf{4.32} & \bf{2.92}\\

\midrule
\textbf{\textsc{S2M2+Lin.Classif}} & WRN & 4.93 & 2.31 & 33.23 & 36.84 \\
\textbf{\textsc{SimpleShot}} & WRN & 4.05 & 1.80 & 39.56 & 55.68 \\
\textbf{\textsc{QDA}} & WRN & 4.52 & 1.78 & 35.95 & 18.53\\ 
\textbf{\textsc{MQDA-MAP}} & WRN & 3.94 &0.94 & 31.17 & 17.37 \\ 
\textbf{\textsc{MQDA-FB}}& WRN & \bf{2.71} & \bf{0.74} & \bf{30.68} & \bf{15.86} \\
\bottomrule
\end{tabular}%
}
\caption{\small \small
\textbf{Calibration error (ECE) comparison on \miniIN.} Lower is better. TS indicates temperature scaling. 
}
\vspace{-0.5em}
\label{tab:calibration}
\end{table}


\cut{
\setlength{\tabcolsep}{4.8pt}
\begin{table}[tb]
\centering
\footnotesize
\begin{tabular}{@{} llcc @{}}
\toprule
{\bf Model}  & {\bf Backbone} & {1-shot ECE} & {5-shot ECE} \\

\midrule 
\textbf{\textsc{NCC}} & Conv-4 & 3.82 & 3.35 \\
\textbf{\textsc{MetaLDA}} & Conv-4 & 2.60 & 0.46 \\ 
\textbf{\textsc{MetaQDA\_MAP}} & Conv-4 & 2.75 & 0.89 \\
\textbf{\textsc{MetaQDA\_FB}}& Conv-4 & \bf{2.33} & \bf{0.45} \\

\midrule

\textbf{\textsc{MAML}} & Conv-4 & 4.04 & 1.46 \\
\textbf{\textsc{TCMAML}} & Conv-4 & 0.80 & 1.62 \\ 
\textbf{\textsc{ProtoNet}} & Conv-4 & 1.26 & 0.70 \\
\textbf{\textsc{TCProtoNet}}& Conv-4 & 0.89 & 0.55 \\

\midrule

\textbf{\textsc{NCC}} & WRN & - & - \\
\textbf{\textsc{MetaLDA}} & WRN & 2.86 & 1.05 \\ 
\textbf{\textsc{MetaQDA\_MAP}} & WRN & 3.94 & 0.91 \\
\textbf{\textsc{MetaQDA\_FB}}& WRN & - & 0.93 \\

\bottomrule
\end{tabular}%
\caption{\small \small
\textbf{Calibration error comparison on \miniIN.} Lower the better for ECE. Best performance are bold.
}
\label{tab:calibration}
\end{table}

}

\subsection{Further Analysis}

\setlength{\tabcolsep}{4.8pt}
\begin{table}[tb]
\centering
\footnotesize
\begin{tabular}{@{} llcc @{}}
\toprule
{\bf Model} & {\bf Backbone} & 1-shot & 5-shot \\
\midrule 

\textbf{\textsc{LDA}} & Conv-4 & - & 64.24 $\pm$ 1.42\% \\
\textbf{\textsc{QDA}} & Conv-4 & - & 34.45 $\pm$ 0.67\% \\
\textbf{\textsc{LDA (prior)}} & Conv-4 & 54.84 $\pm$ 0.80\% & 71.48 $\pm$ 0.64\% \\ 
\textbf{\textsc{QDA (prior)}} & Conv-4 & 54.84 $\pm$ 0.80\% & 71.40 $\pm$ 0.64\% \\
\textbf{\textsc{MetaLDA}} & Conv-4 & 56.24 $\pm$ 0.80\% & 72.39 $\pm$ 0.64\% \\ 
\textbf{\textsc{MetaQDA}} & Conv-4 & \bf{56.41 $\pm$ 0.80\%} & \bf{72.64 $\pm$ 0.62\%} \\ 
\midrule 
\textbf{\textsc{LDA}} & WRN & - & 51.83 $\pm$ 1.29\% \\ 
\textbf{\textsc{QDA}} & WRN & - & 27.14 $\pm$ 0.59\% \\
\textbf{\textsc{LDA (prior)}} & WRN & 63.79 $\pm$ 0.83\% & 81.05 $\pm$ 0.56\% \\ 
\textbf{\textsc{QDA (prior)}} & WRN & 63.79 $\pm$ 0.83\% & 81.18 $\pm$ 0.56\% \\ 
\textbf{\textsc{MetaLDA}} & WRN & 64.92 $\pm$ 0.85\% & 83.18 $\pm$ 0.83\% \\ 
\textbf{\textsc{MetaQDA}} & WRN & \bf{67.83 $\pm$ 0.64\%} & \bf{84.28 $\pm$ 0.69\%} \\ 

\bottomrule
\end{tabular}%
\caption{\small \small
\textbf{Comparison of different classifiers and hand-crafted vs. meta-learned prior measured on \miniIN. }
}
\label{tab:ablation1}
\end{table}

\keypoint{Discussion: Why QDA but not other classifiers?}
In principle, one could attempt an analogous Bayesian meta-learning approach to other classifiers, but we build on discriminant analysis. This is because most classifiers do not admit a tractable Bayesian treatment, besides logistic regression (LR) and discriminant analysis. While LR has a Bayesian generalization \cite{mackay1992bayesLR}, it requires approximate inference and is significantly more complicated to implement, making it difficult to extend to meta-learning. In contrast our generative discriminant analysis approach admits an exact closed form solution, and is easy to extend to meta-learning.


\keypoint{Comparison of Discriminant Analysis Methods}
We compare how moving from LDA to QDA changes performance; and study the impact of changing from (i) no prior, (ii) hand-crafted NIW prior, and (iii) meta-learned prior. We set the hard-crafted NIW prior to $\vec m=0$, $\kappa=1$, $S=I$, and $\nu=d$ which worked well in practice. Table~\ref{tab:ablation1} demonstrates that classic unregularized discriminant analysis methods (LDA and QDA without priors) have very poor performance in the few-shot setting, due to extreme overfitting. This can be seen because: 1) the higher capacity QDA exhibits worse performance than the lower capacity LDA; and 2) incorporating a prior into LDA and QDA, thereby reducing model capacity and overfitting, results in an improvement in performance. Finally, by meta-learning the prior, we are able to optimize inductive bias for few-shot learning performance. Both LDA and QDA benefit from meta-learning, but QDA performs better overall.

\cut{
\subsubsection{MAP-Based MetaQDA vs. FB MetaQDA}

\todo{prefer to remove this part, and add the FB calibration to "EXP\_Calibration Section"}

We compare the performance of the two different variants of MetaQDA proposed in Section~\ref{sec:method}. We compare the accuracy and calibration performance of the two methods on 5-way 1- and 5-shot learning on \miniIN. The results are given in Table \ref{tab:ablation2}. 
From here we can see that accuracy difference is within the statistical margin of error, but the calibration of the fully Bayesian approach is significantly better. This is attributable to exact computation of all the integrals due to repeated use of conjugacy. 

\setlength{\tabcolsep}{4.8pt}
\begin{table}[t]
\centering
\resizebox{1.0\columnwidth}{!}{%
\begin{tabular}{@{} lcc|cc @{}}
\toprule
\multirow{2}{*}{\bf MetaQDA Variant} & \multicolumn{2}{c}{1-shot} & \multicolumn{2}{c}{5-shot}  \\
& Accuracy & ECE & Accuracy & ECE \\
\midrule 

\textbf{\textsc{MetaQDA (MAP)}}  & 56.32 $\pm$ 0.79\% & 8.03  &72.54 $\pm$ 0.65\% & 5.27\\
\textbf{\textsc{MetaQDA (FB)}} & 56.45 $\pm$ 0.79\% & 2.33  & 72.32 $\pm$ 0.68\% &  0.92\\

\bottomrule
\end{tabular}%
}
\caption{\small \small
Comparison of maximum a posteriori (MAP) and fully Bayesian (FB) instantiations of MetaQDA on \textit{mini}Imagenet Conv-4.}

\label{tab:ablation2}
\end{table}
}

\keypoint{Non-Bayesian Meta-Learning?} To disentangle the impact of Bayesian modeling from our classifier architecture and episodic meta-learning procedure, we evaluate a non-Bayesian MetaQDA as implemented by performing MAML learning on the  initialization of the QDA covariance factor $L$ (Eq~\ref{eq:decomp}). From Table~\ref{tab:maml} we can see that MAML is worse than MetaQDA in both accuracy and calibration.

\setlength{\tabcolsep}{4.8pt}
\begin{table}[h]
\centering
\footnotesize
\resizebox{1.0\columnwidth}{!}{
\begin{tabular}{@{} llcccc @{}}
\toprule
\bf{Meta Alg.} & \textbf{Backbone} & \textbf{1-shot Acc.} & \textbf{ECE} & \textbf{5-shot Acc.} & \textbf{ECE} \\
\midrule 
MAML & Conv-4 & 54.33 $\pm$ 0.78\% & 52.75 & 69.17 $\pm$ 0.77\%  &  38.84\\
Bayesian & Conv-4 & 56.41 $\pm$ 0.80\% & 8.03 & 72.64 $\pm$ 0.62\%  & 5.27 \\
MAML& ResNet-18 & 63.66 $\pm$ 0.80\% & 58.11 & 77.82 $\pm$ 0.62\% & 44.62 \\
Bayesian& ResNet-18 & 65.12 $\pm$ 0.66\% & 33.56 & 80.98 $\pm$ 0.75\% & 13.86 \\
\bottomrule
\end{tabular}}
\caption{\small \small 
\bf{Comparison of Bayesian vs. non-Bayesian (MAML-based) realisaton of MetaQDA on \miniIN.}
}
\vspace{-1.5em}
\label{tab:maml}
\end{table}


\section{Conclusion}
We propose an efficient shallow meta-learner for few-shot learning. MetaQDA provides a fast exact inference strategy for amortized Bayesian meta-learning through conjugacy, and highlights a distinct avenue of meta-learning research in contrast to meta representation learning. 
The empirical performance of our model exceeds that of others that rely on off-the-shelf feature extractors, and often outperforms those that train extractors specialised for few-shot learning. In particular it excels in a number of challenging but highly practically important metrics including cross-domain few-shot learning, class incremental few-shot learning, and providing accurate probability calibration---a vital property for many applications where safety or reliability is of paramount concern.

\keypoint{Acknowledgements} This work was supported by the Engineering and Physical Sciences Research Council of the UK (EPSRC) Grant number EP/S000631/1 and the UK MOD University Defence Research Collaboration (UDRC) in Signal Processing, and EPSRC Grant EP/R026173/1.

{\small
\bibliographystyle{ieee_fullname}
\bibliography{metaqda}

\begin{thebibliography}{10}\itemsep=-1pt

\bibitem{bateni2020improved}
Peyman Bateni, Raghav Goyal, Vaden Masrani, Frank Wood, and Leonid Sigal.
\newblock Improved few-shot visual classification.
\newblock In {\em CVPR}, 2020.

\bibitem{baxter2000model}
Jonathan Baxter.
\newblock A model of inductive bias learning.
\newblock {\em Journal of Artificial Intelligence Research}, 12:149--198, 2000.

\bibitem{bertinetto2019R2D2}
Luca Bertinetto, Joao~F Henriques, Philip~HS Torr, and Andrea Vedaldi.
\newblock Meta-learning with differentiable closed-form solvers.
\newblock In {\em ICLR}, 2019.

\bibitem{chen2019closerfewshot}
Wei-Yu Chen, Yen-Cheng Liu, Zsolt Kira, Yu-Chiang Wang, and Jia-Bin Huang.
\newblock A closer look at few-shot classification.
\newblock In {\em ICLR}, 2019.

\bibitem{Du2019FairnessPerspective}
M. {Du}, F. {Yang}, N. {Zou}, and X. {Hu}.
\newblock Fairness in deep learning: A computational perspective.
\newblock {\em IEEE Intelligent Systems}, 2020.

\bibitem{dvornik2019dcem}
Nikita Dvornik, Cordelia Schmid, and Julien Mairal.
\newblock Diversity with cooperation: Ensemble methods for few-shot
  classification.
\newblock In {\em ICCV}, 2019.

\bibitem{dvornik2020selecting}
Nikita Dvornik, Cordelia Schmid, and Julien Mairal.
\newblock Selecting relevant features from a multi-domain representation for
  few-shot classification.
\newblock In {\em ECCV}, 2020.

\bibitem{finn2017model}
Chelsea Finn, Pieter Abbeel, and Sergey Levine.
\newblock Model-agnostic meta-learning for fast adaptation of deep networks.
\newblock In {\em ICML}, 2017.

\bibitem{friedman2001elements}
Jerome Friedman, Trevor Hastie, and Robert Tibshirani.
\newblock {\em The elements of statistical learning}, volume~1.
\newblock Springer series in statistics New York, 2001.

\bibitem{garcia2017few}
Victor Garcia and Joan Bruna.
\newblock Few-shot learning with graph neural networks.
\newblock In {\em ICLR}, 2018.

\bibitem{garnelo2018cnp}
Marta Garnelo, Dan Rosenbaum, Christopher Maddison, Tiago Ramalho, David
  Saxton, Murray Shanahan, Yee~Whye Teh, Danilo Rezende, and S.~M.~Ali Eslami.
\newblock Conditional neural processes.
\newblock In {\em ICML}, 2018.

\bibitem{gelman2003bda}
Andrew Gelman, John~B. Carlin, Hal~S. Stern, and Donald~B. Rubin.
\newblock {\em Bayesian Data Analysis}.
\newblock Texts in statistical science. Chapman \& Hall / CRC, 2nd edition,
  2003.

\bibitem{gidaris2018dynamic}
Spyros Gidaris and Nikos Komodakis.
\newblock Dynamic few-shot visual learning without forgetting.
\newblock In {\em CVPR}, 2018.

\bibitem{goldblum2020unraveling}
Micah Goldblum, Steven Reich, Liam Fowl, Renkun Ni, Valeriia Cherepanova, and
  Tom Goldstein.
\newblock Unraveling meta-learning: Understanding feature representations for
  few-shot tasks.
\newblock In {\em ICML}, 2020.

\bibitem{gordon2019metaPred}
Jonathan Gordon, John Bronskill, Matthias Bauer, Sebastian Nowozin, and Richard
  Turner.
\newblock Meta-learning probabilistic inference for prediction.
\newblock In {\em ICLR}, 2019.

\bibitem{grant2018bayesMAML}
Erin Grant, Chelsea Finn, Sergey Levine, Trevor Darrell, and Tom Griffiths.
\newblock Recasting gradient-based meta-learning as hierarchical bayes.
\newblock In {\em ICLR}, 2018.

\bibitem{guo2017calibration}
Chuan Guo, Geoff Pleiss, Yu Sun, and Kilian~Q Weinberger.
\newblock On calibration of modern neural networks.
\newblock In {\em ICML}, 2017.

\bibitem{guo2020boarder}
Yunhui Guo, Noel C.~F. Codella, Leonid Karlinsky, John~R. Smith, Tajana Rosing,
  and Rogerio Feris.
\newblock A broader study of cross-domain few-shot learning.
\newblock In {\em CVPR workshops}, 2020.

\bibitem{hastie2009elements}
Trevor Hastie, Robert Tibshirani, and Jerome Friedman.
\newblock {\em The Elements of Statistical Learning: Data Mining, Inference,
  and Prediction}.
\newblock Springer Science \& Business Media, 2009.

\bibitem{he2016deep}
Kaiming He, Xiangyu Zhang, Shaoqing Ren, and Jian Sun.
\newblock Deep residual learning for image recognition.
\newblock In {\em CVPR}, 2016.

\bibitem{heskes2000empirical}
Tom Heskes.
\newblock Empirical bayes for learning to learn.
\newblock In {\em ICML}, 2000.

\bibitem{hilliard2018few}
Nathan Hilliard, Lawrence Phillips, Scott Howland, Art{\"e}m Yankov, Courtney~D
  Corley, and Nathan~O Hodas.
\newblock Few-shot learning with metric-agnostic conditional embeddings.
\newblock {\em arXiv preprint arXiv:1802.04376}, 2018.

\bibitem{hospedales2020metaSurvey}
Timothy Hospedales, Antreas Antoniou, Paul Micaelli, and Amos Storkey.
\newblock Meta-learning in neural networks: A survey.
\newblock {\em arXiv preprint arXiv:2004.05439}, 2020.

\bibitem{ignatov2019aiSmartphone}
Andrey Ignatov, Radu Timofte, Andrei Kulik, Seungsoo Yang, Ke Wang, Felix Baum,
  Max Wu, Lirong Xu, and Luc Van~Gool.
\newblock Ai benchmark: All about deep learning on smartphones in 2019.
\newblock {\em arXiv preprint arXiv:1910.06663}, 2019.

\bibitem{jiang2019CAML}
Xiang Jiang, Mohammad Havaei, Farshid Varno, Gabriel Chartrand, Nicolas
  Chapados, and Stan Matwin.
\newblock Learning to learn with conditional class dependencies.
\newblock In {\em ICLR}, 2019.

\bibitem{krause2013cars}
Jonathan Krause, Michael Stark, Jia Deng, and Li Fei-Fei.
\newblock 3d object representations for fine-grained categorization.
\newblock In {\em ICCV workshops}, 2013.

\bibitem{krizhevsky2009learning}
Alex Krizhevsky, Geoffrey Hinton, et~al.
\newblock Learning multiple layers of features from tiny images.
\newblock {\em Technical Report}, 2009.

\bibitem{Kuper2018TowardNetworks}
Lindsey Kuper, Guy Katz, Justin Gottschlich, Kyle Julian, Clark Barrett, and
  Mykel Kochenderfer.
\newblock {Toward Scalable Verification for Safety-Critical Deep Networks}.
\newblock In {\em SysML}, 2018.

\bibitem{lecun2010mnist}
Y Lecun and C Cortes.
\newblock Mnist handwritten digit database.
\newblock 2010.

\bibitem{lee2019meta}
Kwonjoon Lee, Subhransu Maji, Avinash Ravichandran, and Stefano Soatto.
\newblock Meta-learning with differentiable convex optimization.
\newblock In {\em CVPR}, 2019.

\bibitem{li2019GCR}
Aoxue Li, Tiange Luo, Tao Xiang, Weiran Huang, and Liwei Wang.
\newblock Few-shot learning with global class representations.
\newblock In {\em ICCV}, 2019.

\bibitem{li2020afhn}
Kai Li, Yulun Zhang, Kunpeng Li, and Yun Fu.
\newblock Adversarial feature hallucination networks for few-shot learning.
\newblock In {\em CVPR}, 2020.

\bibitem{li2017meta}
Zhenguo Li, Fengwei Zhou, Fei Chen, and Hang Li.
\newblock Meta-sgd: Learning to learn quickly for few-shot learning.
\newblock {\em arXiv preprint arXiv:1707.09835}, 2017.

\bibitem{liu2020universal}
Lu Liu, William Hamilton, Guodong Long, Jing Jiang, and Hugo Larochelle.
\newblock A universal representation transformer layer for few-shot image
  classification.
\newblock In {\em ICLR}, 2021.

\bibitem{liu2018transductive}
Yanbin Liu, Juho Lee, Minseop Park, Saehoon Kim, and Yi Yang.
\newblock Transductive propagation network for few-shot learning.
\newblock In {\em ICLR}, 2019.

\bibitem{mackay1992bayesLR}
D.~J.~C. {MacKay}.
\newblock The evidence framework applied to classification networks.
\newblock {\em Neural Computation}, 4(5):720--736, 1992.

\bibitem{mangla2020charting}
Puneet Mangla, Nupur Kumari, Abhishek Sinha, Mayank Singh, Balaji
  Krishnamurthy, and Vineeth~N Balasubramanian.
\newblock Charting the right manifold: Manifold mixup for few-shot learning.
\newblock In {\em WACV}, 2020.

\bibitem{mishra2018simple}
Nikhil Mishra, Mostafa Rohaninejad, Xi Chen, and Pieter Abbeel.
\newblock A simple neural attentive meta-learner.
\newblock In {\em ICLR}, 2018.

\bibitem{murphy2012machine}
Kevin~P Murphy.
\newblock {\em Machine learning: a probabilistic perspective}.
\newblock MIT press, 2012.

\bibitem{nixon2019measuring}
Jeremy Nixon, Michael~W Dusenberry, Linchuan Zhang, Ghassen Jerfel, and Dustin
  Tran.
\newblock Measuring calibration in deep learning.
\newblock In {\em CVPR Workshops}, 2019.

\bibitem{oreshkin2018tadam}
Boris Oreshkin, Pau~Rodr{\'\i}guez L{\'o}pez, and Alexandre Lacoste.
\newblock Tadam: Task dependent adaptive metric for improved few-shot learning.
\newblock In {\em NeurIPS}, 2018.

\bibitem{patacchiola2020bayesian}
Massimiliano Patacchiola, Jack Turner, Elliot~J Crowley, Michael O'Boyle, and
  Amos~J Storkey.
\newblock Bayesian meta-learning for the few-shot setting via deep kernels.
\newblock In {\em NeurIPS}, 2020.

\bibitem{qiao2017few}
Siyuan Qiao, Chenxi Liu, Wei Shen, and Alan~L Yuille.
\newblock Few-shot image recognition by predicting parameters from activations.
\newblock In {\em CVPR}, 2018.

\bibitem{ravi2017optimization}
Sachin Ravi and Hugo Larochelle.
\newblock Optimization as a model for few-shot learning.
\newblock In {\em ICLR}, 2017.

\bibitem{rebuffi2017icarl}
Sylvestre-Alvise Rebuffi, Alexander Kolesnikov, Georg Sperl, and Christoph~H.
  Lampert.
\newblock {iCaRL}: Incremental classifier and representation learning.
\newblock In {\em CVPR}, 2017.

\bibitem{ren2019attentionAttractor}
Mengye Ren, Renjie Liao, Ethan Fetaya, and Richard~S. Zemel.
\newblock Incremental few-shot learning with attention attractor networks.
\newblock {\em NeurIPS}, 2019.

\bibitem{ren2018meta}
Mengye Ren, Eleni Triantafillou, Sachin Ravi, Jake Snell, Kevin Swersky,
  Joshua~B Tenenbaum, Hugo Larochelle, and Richard~S Zemel.
\newblock Meta-learning for semi-supervised few-shot classification.
\newblock In {\em ICLR}, 2018.

\bibitem{requeima2019cnaps}
James Requeima, Jonathan Gordon, John Bronskill, Sebastian Nowozin, and
  Richard~E Turner.
\newblock Fast and flexible multi-task classification using conditional neural
  adaptive processes.
\newblock In {\em NIPS}, 2019.

\bibitem{russakovsky2015imagenet}
Olga Russakovsky, Jia Deng, Hao Su, Jonathan Krause, Sanjeev Satheesh, Sean Ma,
  Zhiheng Huang, Andrej Karpathy, Aditya Khosla, Michael Bernstein, et~al.
\newblock Imagenet large scale visual recognition challenge.
\newblock {\em IJCV}, 115(3):211--252, 2015.

\bibitem{rusu2019leo}
Andrei~A Rusu, Dushyant Rao, Jakub Sygnowski, Oriol Vinyals, Razvan Pascanu,
  Simon Osindero, and Raia Hadsell.
\newblock Meta-learning with latent embedding optimization.
\newblock In {\em ICLR}, 2019.

\bibitem{santoro2017simple}
Adam Santoro, David Raposo, David~GT Barrett, Mateusz Malinowski, Razvan
  Pascanu, Peter Battaglia, and Timothy Lillicrap.
\newblock A simple neural network module for relational reasoning.
\newblock In {\em NeurIPS}, 2017.

\bibitem{snell2017prototypical}
Jake Snell, Kevin Swersky, and Richard~S Zemel.
\newblock Prototypical networks for few-shot learning.
\newblock In {\em NeurIPS}, 2017.

\bibitem{sun2020explanation}
Jiamei Sun, Sebastian Lapuschkin, Wojciech Samek, Yunqing Zhao, Ngai-Man
  Cheung, and Alexander Binder.
\newblock Explanation-guided training for cross-domain few-shot classification.
\newblock {\em arXiv preprint arXiv:2007.08790}, 2020.

\bibitem{sun2019meta}
Qianru Sun, Yaoyao Liu, Tat-Seng Chua, and Bernt Schiele.
\newblock Meta-transfer learning for few-shot learning.
\newblock In {\em CVPR}, 2019.

\bibitem{sung2018learning}
Flood Sung, Yongxin Yang, Li Zhang, Tao Xiang, Philip~HS Torr, and Timothy~M
  Hospedales.
\newblock Learning to compare: Relation network for few-shot learning.
\newblock In {\em CVPR}, 2018.

\bibitem{tao2020few}
Xiaoyu Tao, Xiaopeng Hong, Xinyuan Chang, Songlin Dong, Xing Wei, and Yihong
  Gong.
\newblock Few-shot class-incremental learning.
\newblock In {\em CVPR}, 2020.

\bibitem{triantafillou2019meta}
Eleni Triantafillou, Tyler Zhu, Vincent Dumoulin, Pascal Lamblin, Evci, et~al.
\newblock Meta-dataset: A dataset of datasets for learning to learn from few
  examples.
\newblock In {\em ICLR}, 2019.

\bibitem{tseng2020cross}
Hung-Yu Tseng, Hsin-Ying Lee, Jia-Bin Huang, and Ming-Hsuan Yang.
\newblock Cross-domain few-shot classification via learned feature-wise
  transformation.
\newblock In {\em ICLR}, 2020.

\bibitem{vinyals2016matching}
Oriol Vinyals, Charles Blundell, Tim Lillicrap, Daan Wierstra, et~al.
\newblock Matching networks for one shot learning.
\newblock In {\em NeurIPS}, 2016.

\bibitem{wang2020comparison}
Hongyu Wang, Henry Gouk, Eibe Frank, Bernhard Pfahringer, and Michael Mayo.
\newblock A comparison of machine learning methods for cross-domain few-shot
  learning.
\newblock In {\em AJCAI}, 2020.

\bibitem{wang2019simpleshot}
Yan Wang, Wei-Lun Chao, Kilian~Q Weinberger, and Laurens van~der Maaten.
\newblock Simpleshot: Revisiting nearest-neighbor classification for few-shot
  learning.
\newblock {\em arXiv preprint arXiv:1911.04623}, 2019.

\bibitem{wang2019fewShotSurvey}
Yaqing Wang, Quanming Yao, James~T. Kwok, and Lionel~M. Ni.
\newblock Generalizing from a few examples: A survey on few-shot learning.
\newblock {\em ACM Comput. Surv.}, 53(3), June 2020.

\bibitem{xing2019am3}
Chen Xing, Negar Rostamzadeh, Boris Oreshkin, and Pedro~O O~Pinheiro.
\newblock Adaptive cross-modal few-shot learning.
\newblock {\em NeurIPS}, 2019.

\bibitem{yin2020metaMemorisation}
Mingzhang Yin, George Tucker, Mingyuan Zhou, Sergey Levine, and Chelsea Finn.
\newblock Meta-learning without memorization.
\newblock {\em ICLR}, 2020.

\bibitem{yoon2018bayesian}
Jaesik Yoon, Taesup Kim, Ousmane Dia, Sungwoong Kim, Yoshua Bengio, and Sungjin
  Ahn.
\newblock Bayesian model-agnostic meta-learning.
\newblock In {\em NeurIPS}, 2018.

\bibitem{yoon2019tapnet}
Sung~Whan Yoon, Jun Seo, and Jaekyun Moon.
\newblock Tapnet: Neural network augmented with task-adaptive projection for
  few-shot learning.
\newblock In {\em ICML}, 2019.

\bibitem{yosinski2014howTransferable}
Jason Yosinski, Jeff Clune, Yoshua Bengio, and Hod Lipson.
\newblock How transferable are features in deep neural networks?
\newblock In {\em NeurIPS}, 2014.

\bibitem{xueting2020dcn}
Xueting Zhang, Yuting Qiang, Sung Flood, Yongxin Yang, and Timothy~M.
  Hospedales.
\newblock {RelationNet2:} deep comparison columns for few-shot learning.
\newblock In {\em IJCNN}, 2020.

\bibitem{zintgraf2018cavia}
Luisa~M Zintgraf, Kyriacos Shiarlis, Vitaly Kurin, Katja Hofmann, and Shimon
  Whiteson.
\newblock Fast context adaptation via meta-learning.
\newblock In {\em ICML}, 2019.

\end{thebibliography}
}

\appendix
\newpage

\section{Illustrative Schematic of MetaQDA}
\begin{figure*}[!h]
\centering
\subcaptionbox{\label{a}}{\includegraphics[width = .45\linewidth]{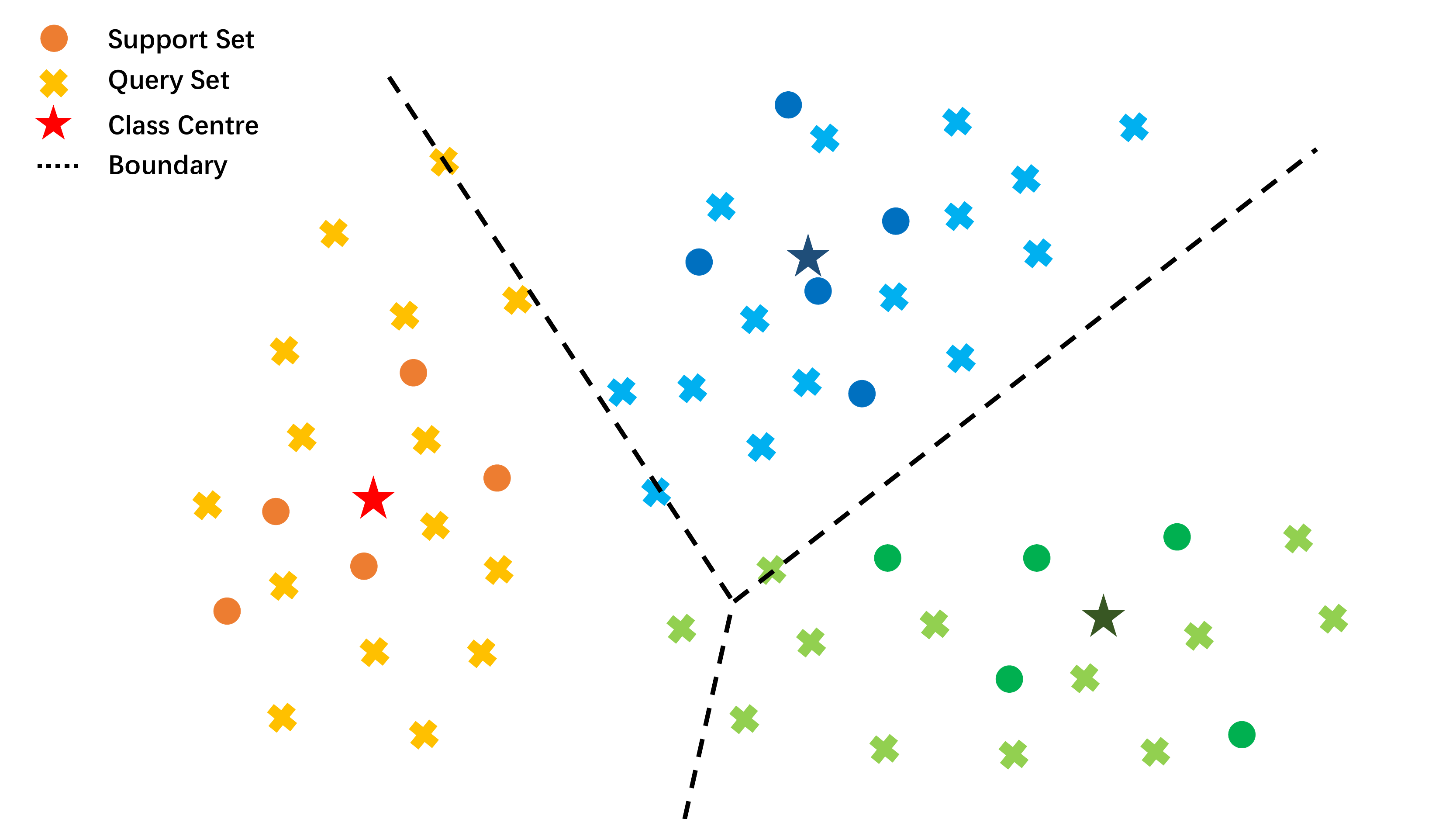}}
\subcaptionbox{\label{b}}{\includegraphics[width = .45\linewidth]{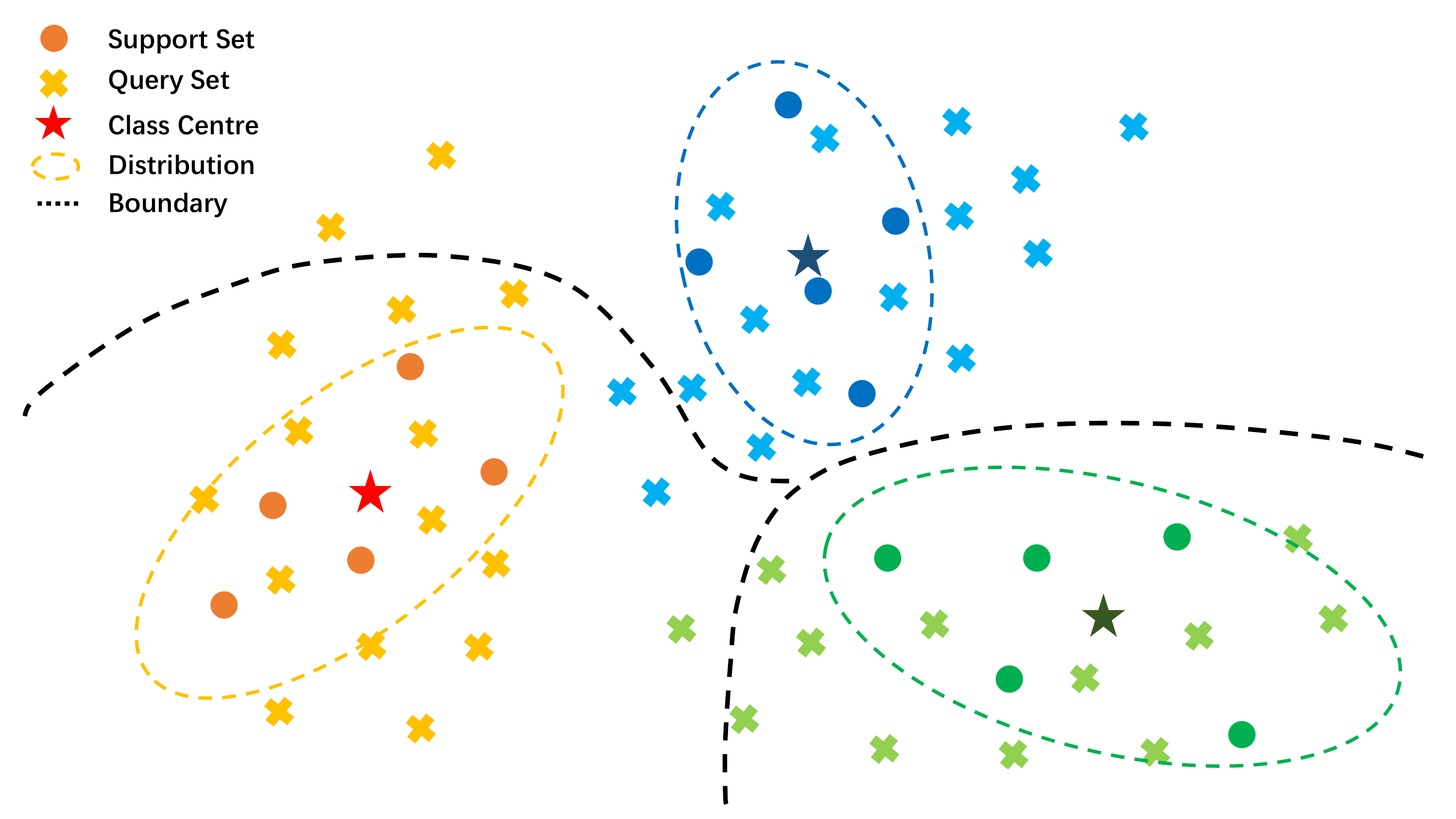}}
\\
\subcaptionbox{\label{c1}}{\includegraphics[width = .9\linewidth]{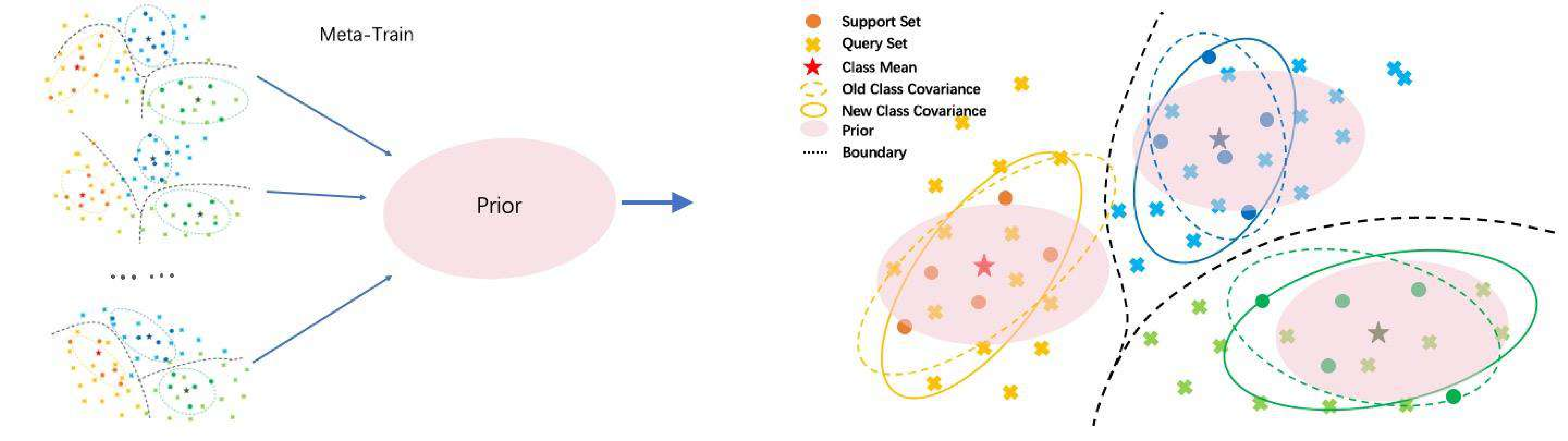}}

\caption{
\small \small
\textbf{Illustrative Schematic of MetaQDA.}
(a) NCC classifier uses the class mean to induce linear decision boundaries. (b) QDA uses both the support class mean and covariance to induce a curved decision boundary, but easily overfits in a few-shot regime due. (c) MetaQDA meta-learns the QDA parameter prior to provide stable estimation of a non-linear decision boundary without overfitting.}
\label{fig:visualization}
\vskip 0.5cm
\end{figure*}

To illustrate the mechanism of MetaQDA, we compare it schematically to conventional linear classifier used in many studies \cite{chen2019closerfewshot,snell2017prototypical,liu2020universal}, and vanilla QDA in Figure~\ref{fig:visualization}. In the figure, the colored circles indicate 3-way-5-shot support datasets, and the "x" data points with are the query set of the corresponding color. The dashed line is the decision boundary of different classifiers.
Figure~\ref{fig:visualization}(a) shows Nearest Centre Classifier (NCC) \cite{snell2017prototypical,liu2020universal}, where the stars represents the mean of the support set class distributions, and these induce linear decision boundaries. Figure~\ref{fig:visualization}(b) depicts the Quadratic Discriminant Analysis (QDA) classifier, where the dashed ellipses represents the class covariance models, estimated from the support set. These induce a non-linear decision boundary. Figure~\ref{fig:visualization}(c) illustrates our MetaQDA, where the meta-training process learns a shared NIW prior (the shadow ellipse) from many few-shot training tasks. Then MetaQDA uses conjugacy to update the class covariances (solid line) using the support set and prior, and so induces a better non-linear decision boundary. 

This illustrates how the MetaQDA setup allows us to exploit the benefit of a non-linear classifier, without the associated overfitting risk that would normally undermine such an attempt (as illustrated by the poor results of vanilla MetaQDA in Tab 7, 8 of the main manuscript).

\section{Additional Experimental Setting Details: Standard Few-shot Learning}\label{sec:extraHypers}

\keypoint{Parameters for training the Conv-4 extractor} 
Following \cite{wang2019simpleshot}, we use stochastic gradient descent (SGD) with a multi-step learning rate schedule, momentum of 0.9, and the initial learning rate is set to 0.01 for both {\miniIN} and CIFAR-FS, and 0.001 for \tierIN. At epochs 70 and 100 we reduce the learning rate by a factor of 0.1. Weight decay is set as 0.0001 through
out training. 

\keypoint{Parameters for training the ResNet-18 extractor} 
Following \cite{wang2019simpleshot}, we use stochastic gradient descent (SGD) with a multi-step learning rate schedule, momentum of 0.9, and the initial learning rate is set to 0.001 for \tierIN. At epochs 70 and 100 we reduce the learning rate by a factor of 0.1. Weight decay is set as 0.0001 throughout training. Batch size is 256 images.

\keypoint{Parameters for training the WRN-28-10 extractor} 
Following \cite{mangla2020charting}, as for 1-shot classification on \miniIN, we use stochastic gradient descent (SGD) with a multi-step learning rate schedule, momentum of 0.9, and the initial learning rate is set to 0.001. For 5-shot classification on \miniIN and 1-shot classification on \tierIN, we use ADAM optimiser. For CIFAR-FS, we use the pre-trained WRN backbone of S2M2.

\section{Additional Experimental Setting Details: Few-Shot Class Incremental Learning}\label{sec:extraHypers}

\keypoint{Training setup} We follow the experimental setup of \cite{tao2020few}. Specifically, we use the same 60 base classes to pre-train an initial ResNet-18 backbone using mini-batch size as 128 and use stochastic gradient descent (SGD) with the initial learning rate of 0.1, decreasing the learning rate to 0.01/0.001 after 30/40 epochs, respectively.

Meta-Training: The MetaQDA prior is then trained using Algorithm 1 (main manuscript) by generating episodes from the 60 base class set, using the feature extractor trained as above. 

Meta-Testing: Due to our Bayesian class-conditional modeling, meta-testing decomposes over classes. Class-incremental learning is thus trivially realized by running MetaQDA's update step for each new category, and adding the final mean and covariance to the set used by the final QDA classifier. We apply MetaQDA both for the many-shot base classes, and 5-shot incrementally added classes.

The results in Tab 6 of the main manuscript are averages generated by independently repeating both meta-train and meta-test (8 incremental sessions each) phases 10 times.

\section{Full Meta-Dataset Results}\label{sec:fullMetaDataset}

\keypoint{Implementation Details}
We use the same backbone as SUR \cite{dvornik2020selecting} and URT \cite{liu2020universal}, and take the trained fused features by URT \cite{liu2020universal}.
We use ADAM optimizer and cosine learning rate scheduler, and the initial learning rate is set to 0.0003, beta is set as 0.9 and 0.999. Weight decay is set as 0.0001 throughout training. The number of training episodes is 10000.

\keypoint{Results}
Following \cite{triantafillou2019meta}, few-shot tasks are sampled with varing number of classes $N$, varying number of shots $K$ and class imbalance. Table~\ref{tab:fullmetadataset} reports performance in accuracy over  over 600 sampled meta-test tasks. Because most of the results have very similar confidence interval, we omit this part to make the table more readable.
The results of other SotA algorithms are taken from URT \cite{liu2020universal} and SCNAP\cite{bateni2020improved}. From the results we can see that MetaQDA performs well in both seen domains (left) and out-of-distribution unseen (right) domains. It achieves highest performance in 8 of 13 domains within the meta-dataset benchmark.

\setlength{\tabcolsep}{4.6pt}
\begin{table*}[h]
\centering
\footnotesize
\begin{tabular}{@{} l|cccccccc|ccccc @{}}
\toprule

{\bf Model} & ImageNet & Omniglot & Aircraft & Birds & DTD & Quickdraw & Fungi & Flower & Signs & Mscoco & MNIST & CIFAR10 & CIFAR100 \\
\midrule 

\textbf{\textsc{MAML}} \cite{finn2017model} & 32.4 & 71.9 & 52.8 & 47.2 & 56.7 & 50.5 & 21.0 & 70.9 & 34.2 & 24.1 & NA & NA & NA\\
\textbf{\textsc{RelationNet}} \cite{sung2018learning} & 30.9 & 86.6 & 69.7 & 54.1 & 56.6 & 61.8 & 32.6 & 76.1 & 37.5 & 27.4 & NA & NA & NA  \\
\textbf{\textsc{MatchingNet}} \cite{vinyals2016matching} & 36.1 & 78.3 & 69.2 & 56.4 & 61.8 & 60.8 & 33.7 & 81.9 & 55.6 & 28.8 & NA & NA & NA  \\
\textbf{\textsc{Finetune}} \cite{yosinski2014howTransferable} & 43.1 & 71.1 & 72.0 & 59.8 & 69.1 & 47.1 & 38.2 & 85.3 & 66.7 & 35.2 & NA & NA & NA  \\
\textbf{\textsc{ProtoNet}} \cite{snell2017prototypical} & 44.5 & 79.6 & 71.1 & 67.0 & 65.2 & 64.9 & 40.3 & 86.9 & 46.5 & 39.9 & 74.3 & 66.4 & 54.7 \\
\textbf{\textsc{CNAP}} \cite{requeima2019cnaps}& 51.3 & 88.0 & 76.8 & 71.4 & 62.5 & 71.9 & 46.0 & 89.2 & 60.1 & 42.3 & 88.6 & 60.0 & 48.1 \\
\textbf{\textsc{SCNAP}} \cite{bateni2020improved} & \bf{58.6} & 91.7 & 82.4 & 74.9 & 67.8 & 77.7 & 46.9 & \bf{90.7} & 73.5 & 46.2 & 93.9 & 74.3 & 60.5 \\
\textbf{\textsc{SUR}} \cite{dvornik2020selecting}& 56.3 & 93.1 & 85.4 & 71.4 & 71.5 & 81.3 & 63.1 & 82.8 & 70.4 & \bf{52.4} & 94.3 & 66.8 & 56.6 \\
\textbf{\textsc{URT}} \cite{liu2020universal} & 55.7 & 94.9 & 85.8 & \bf{76.3} & 71.8 & 82.5 & 63.5 & 88.2 & 69.4 & 52.2 & \bf{94.8} & 67.3 & 56.9 \\
\midrule

\textbf{\textsc{MetaQDA}} & 56.5 & \bf{96.3} & \bf{86.5} & 75.1 & \bf{73.4} & \bf{82.6} & \bf{63.7} & 87.4 & \bf{73.8} & 49.8 & 94.3 & \bf{68.2} & \bf{57.8} \\

\bottomrule
\end{tabular}%
\caption{\small \small
\textbf{Full details of testing performance on the extended meta-dataset benchmark.} Left is the in-domain (seen) dataset performance, where MetaQDA ranks first 5 times in 8 domains. Right is the out-of-domain (unseen) dataset performance, where MetaQDA ranks first 3 times in 5 domains. Overall, MetaQDA has state-of-the-art performance.
}
\vskip 0.5cm
\label{tab:fullmetadataset}
\end{table*}

\end{document}